**Title:**
Computational Synthesis of Wearable Robot Mechanisms: Application to Hip-Joint Mechanisms

**Authors:**
S. W. Kang[1]*, J. Ryu[1], S. I. Kim[1]†, Y. Kim[1]‡, Y. Y. Kim[1]§

**Affiliations:**
[1]Department of Mechanical and Aerospace Engineering and Institute of Advanced Machines and Design, Seoul National University, Seoul, South Korea.
*Present address: Department of Civil and Systems Engineering, Johns Hopkins University, Baltimore, MD, USA.
†Present address: Department of Mechanical Engineering, Massachusetts Institute of Technology, Cambridge, MA, USA.
‡Present address: School of Mechanical Engineering, Pusan National University, Busan, South Korea

§Corresponding author. Email: yykim@snu.ac.kr

**Abstract:**
Since wearable linkage mechanisms could control the moment transmission from actuator(s) to wearers, they can help ensure that even low-cost wearable systems provide advanced functionality tailored to users' needs. For example, if a hip mechanism transforms an input torque into a spatially-varying moment, a wearer can get effective assistance both in the sagittal and frontal planes during walking, even with an affordable single-actuator system. However, due to the combinatorial nature of the linkage mechanism design space, the topologies of such nonlinear-moment-generating mechanisms are challenging to determine, even with significant computational resources and numerical data. Furthermore, on-premise production development and interactive design are nearly impossible in conventional synthesis approaches. Here, we propose an innovative autonomous computational approach for synthesizing such wearable robot mechanisms, eliminating the need for exhaustive searches or numerous data sets. Our method transforms the synthesis problem into a gradient-based optimization problem with sophisticated objective and constraint functions while ensuring the desired degree of freedom, range of motion, and force transmission characteristics. To generate arbitrary mechanism topologies and dimensions, we employed a unified ground model. By applying the proposed method for the design of hip joint mechanisms, the topologies and dimensions of non-series-type hip joint mechanisms were obtained. Biomechanical simulations validated its multi-moment assistance capability, and its wearability was verified via prototype fabrication. The proposed design strategy can open a new way to design various wearable robot mechanisms, such as shoulders, knees, and ankles.

**One-Sentence Summary:** We propose an innovative computational method to autonomously synthesize wearable robot mechanisms without needing any baseline designs or numerous data.

**Main Text:**

# INTRODUCTION

Design algorithms for wearable robots that could consider individuals' requirements will become increasingly important as digital manufacturing, such as additive manufacturing, has grown rapidly. The manufacture of sophisticated/customized wearable hardware is becoming affordable, and consumers will be able to access them within a reduced lead time in the foreseeable future (1). Specifically, design algorithms for linkage mechanisms will be one of the most significant concerns in this trend since specialized mechanisms can be considered sophisticated motion and force/moment transmitters providing intelligence to wearable robots (2-4). For example, consumers will be able to access affordable wearable gait-assistive robots producing tailored three-dimensional moment patterns (see, e.g., Fig. S1) generated from sophiscated linkage mechanisms integrated with a single actuator. However, available methods (see, e.g., (5)) for tuning pre-selected wearable robot mechanisms may not yield creative solutions for individuals because they typically start with mechanisms of specific topologies. Besides, it is preferable to conduct the design on a personal computer rather than on a cloud service to avoid any cyber-security issue (6) since sensitive private information, such as human body shapes, kinematic/physiological data, and movement patterns of the wearers, etc., is needed for the design. Therefore, we propose a novel optimization-based design algorithm that enables advanced topology and dimensional synthesis of wearable robot mechanisms even on typical desktop computers; the desired moment transmission characteristics on motions to be assisted are fully reflected. Specifically, our method was developed to computationally synthesize hip mechanisms of gait assistive exo-robot generating *spatially-varying moments only with a single actuator*, yielding unprecedented mechanisms. Furthermore, our algorithm completes the syntheses within tens of minutes on desktop computers. This agile design searching capability will make this algorithm suitable for on-premises product development pipelines and interactive design tools in the future (7, 8).

Computational methods to design robot mechanisms mostly use expensive gradient-free techniques like evolutionary algorithms (9) or stochastic graph search algorithms (10). Recent advancements in learning-based methods have made them more feasible (11, 12). For example, one study was able to design the morphology and controller of robots at the same time by using deep reinforcement learning and heuristic graph search (11), and another was able to design a general 1-DOF (Degree of Freedom) linkage mechanism by training a dataset of 100 million samples (12). While not directly related to the design of linkage mechanisms, a recent study suggested the possibility of utilizing generative AI such as ChatGPT in a robot design process (13). But the studies may not provide creative solutions efficiently in the far outside domain of data used for learning and have considered relatively simple topologies (e.g., tree-like structures or 1-DOF mechanisms) due to the computationally expensive combinatorial nature of design space for linkage mechanism topologies.

In our study, on the other hand, we aim to develop an algorithm allowing an agile exploration of a much broader range of general topologies to generate creative solutions under so-far-unexplored design conditions. For instance, the algorithm should be able to generate even hybrid serial-parallel structures with multiple DOFs. To find creative design solutions efficiently, we propose a gradient-based topology optimization approach. The efficiency of gradient-based methods has been demonstrated in the synthesis of certain link mechanisms composed of revolute joints and links (14-17), various types of joints and links (18), links and gears (19), rigid-links and elastic components (20). They were also applied to design industrial products such as a vehicle suspension (21, 22), a swimming-motion (23) or finger guiding (24) mechanism. But all of these methods were only able to generate "fully actuated" motion/path guiding mechanisms, which have the same number of DOFs as the number of actuators. Meanwhile, wearable assistive robots are often designed as "underactuated" systems, implying that the number of actuators is fewer than the DOF of the mechanism. This is because the robot is not acting as a master and manipulating the wearers but rather is intended to support only a portion of their movements (25). Particularly, no mechanism synthesis method for underactuated mechanisms satisfying torque transmission requirements essential for human assistance has been conceived so far. In this baseline-free synthesis method, we tackle a challenging issue: an underactuated multi-DOF wearable robot design considering input torque transmitting criteria. To this end, we propose a new topology optimization formulation to address the design of torque transmission characteristics in underactuated mechanisms.

Figure 1 shows the flowchart of our proposed computational design framework for wearable robot mechanisms, which will be named "ExoSynthesizer." When given the design domain ( $\Omega$ ), the desired range-of-motion (or workspace), and the desired force/torque transmittance properties between the actuator(s) and the end-effector at each point (or pose) in the workspace, the design algorithm outputs the topology and dimensions of the wearable mechanism that satisfy inputs. Note that no baseline information is given as input.

As examples of design, we focus mostly on hip-exo robot mechanisms. A hip joint moment during walking has spatially-varying direction to provide forward propulsion and lateral balancing (see Fig. S1). Some robots provide direction-varying hip moments by using multiple actuators (26, 27), and others for paraplegics use inconvenient crutches for balancing in addition to forward propulsion assistance from robots (28-31). However, to lower price and weight, most mobile-type gait-assistive hip-exo robots only assist flexion/extension moment with a single-actuator per hip joint (32-36). All of them (26-36) use series-type mechanisms for the hip joint to control moments independently in the frontal, transverse, and sagittal planes, but they use advanced control methods (37–40) to give an optimized or customized magnitude pattern along the fixed direction. On the other hand, we propose a new hip-exo robot concept that can provide a customized hip moment with a coupled direction-varying moment with a single actuator connected to an advanced linkage mechanism. With the concept of a robot, individuals could be assisted both in lateral balancing and/or forward propulsion, even with a single actuator, in a customized way. Since neither such a mechanism nor a method for synthesizing such mechanisms exists in the literature, ExoSynthesizer is essential in synthesizing them. Also, since the relationship between output moment direction and hip angle is fixed once the mechanism design is fixed, the user needs to iteratively and interactively update the design based on their preferences. Therefore, we need an agile and creative synthesis method such as our ExoSynthesizer.

Figuare 2(a) shows the problem definition of the most distinctive problem, Case Study 1, out of all three examples dealt with in this paper. The design domain in green denotes the physical space that a synthesized mechanism can occupy. The input actuation is fixed to the abduction/adduction direction. The mechanism to be synthesized transforms the input torque into direction-varying output moments ( $\mathbf{M}_{t*}^{out,EF}$ ) denoted as large arrows depending on gait phase (or leg poses) ($t* = 1\sim7$). Figure 2(b) shows detailed output moment directions and leg poses (or hip angles). The problem definitions for other case studies are given in Tables S1 and S2. OpenSim will be used to validate the functionality of the mechanism synthesized in Case Study 1 by our approach and the reduction of the biological moment of a person wearing them (41). As shown in Figure 1, a prototype of the synthesized mechanism in Case Study 1 is fabricated to show its wearability and kinematic movements.

To assist the reader in comprehending the autonomous mechanism synthesis approach, which will be described in the "Results" section, we will provide a brief summary of the synthesis flow of our method. According to Figure 1, we first transform our autonomous mechanism synthesis problem into an optimization problem based on real variables. The design domain ($\Omega$) is discretized using a ground model called the spring-connected spherical rigid block model (Spherical-SBM or S-SBM) in order to apply ExoSynthesizer. The SBM (spring-connected rigid block model) used previously (15) was only defined in a plane, so it has been modified as the S-SBM to be applicable to a spherical surface. We use the S-SBM because mechanisms of different topologies and shapes can be presented by the same model (See (15) for the planar SBM instance). Because a human hip joint can be seen as a mechanical ball joint, spherical mechanisms are preferred for wearable hip mechanisms to enable smooth interaction between the wearable hip robot and its wearer. The spherical rigid blocks in the S-SBM are connected by artificial springs of zero length whose stiffness is modeled to vary continuously between their minimum and maximum values.

The novel contribution of our computational synthesis method is in its formulation using a sophisticated objective function. Specifically, we newly define the so-called reverse work transmittance efficiency function to ensure that the synthesized exoskeleton mechanism meets a particular condition on DOFs and achieves the desired range-of-motion. The proposed formulation based on this efficiency is completely distinct from the formulations offered in (14-24); as previous studies were only concerned with fully-actuated mechanisms, whereas the current study must deal with underactuated mechanisms. The intended moment transmittance properties, on the other hand, are regarded as a constraint function. Then, ExoSynthesizer iteratively searches for the desired mechanism using the proposed formulation based on the spherical ground model, as shown in the optimization history at the bottom of Fig. 1. (Red line: objective function value; blue line: constraint function value.) With this summary, we shall introduce the method of autonomous synthesis of wearable robot mechanisms, especially for hip joints.

## RESULTS
### Modeling with spherical-SBM

As stated in the Introduction, hip-exo robot mechanisms will be synthesized solving an optimization problem defined on a ground model called the S-SBM (spring-connected spherical rigid block model) that is capable of representing an arbitrary spherical mechanism (within the discretization degree). The spherical version of the SBM should be used here to ensure that the remote center-of-rotation property of spherical mechanisms can naturally satisfy some intrinsic requirements that a wearable hip mechanism should not interfere with its wearer's body while assisting the spherical movement of the hip joint located inside the wearer's body.

As shown in the 'Modeling' part of Fig. 1 as well as in Fig. 2A, the S-SBM is constructed by discretizing the spherical design surface domain ($\Omega$) into two-dimensional spherical rigid blocks connected at their corners by "artificial" zero-length springs. The symbol *O* in Fig. 2A denotes the origin of the spherical coordinate system that coincides with the center of the right hip joint of the wearer. The XY, YZ, and ZX planes are synchronized with the transverse, sagittal, and frontal planes, respectively. The domain $\Omega$ is a part of the spherical surface defined by $\{(\Theta,\Phi)|(\pi/2-\Theta_T)\leq\Theta\leq\pi/2, -\Phi_P\leq\Phi\leq 0\}$, where $\Theta$ and $\Phi$ are the polar and azimuthal angles, respectively. The symbols $\Phi_P$ and $\Theta_T$ define the ranges of the design surface domain.

Figures 3A and 3B, respectively, show the interconnectivity between two adjacent blocks and the discretized design domains[1] where the interconnecting spring stiffness $k$ is allowed to continuously vary between its minimum $k_{min} \approx 0$ and its maximum $k_{max}$. As shown in Fig. 3A, each pair of adjacent blocks may be connected to each other at their common corners by zero-length elastic springs with the maximum stiffness value, which function as revolute joints. (The disconnected condition is represented by springs with the minimum stiffness value.) In addition, it is assumed that all blocks are connected to the origin of the spherical surface by ball joints. Figure 3B illustrates the representation of different mechanisms using the same S-SBM with the adjusted spring stiffness values ($k_{min} \approx 0$ or $k_{max}$). For instance, the left illustration in Fig. 3B shows that Blocks 1 and 5 as a whole represent Link 1, Blocks 2-3 and 6-8 as a whole represent Link 2, and Block 4 denotes Link 3 if the stiffness values (either $k_{min}$ or $k_{max}$) are properly adjusted. Therefore, the S-SBM with the interconnecting zero-length springs of the adjusted stiffness can represent a series R-R-R mechanism, where 'R' stands for a revolution joint. On the other hand, we can also represent a mechanism with a closed chain (a 4-bar mechanism) on the right side of Figure 3B using the same S-SBM with different sets of the stiffness values. The resulting mechanism will be referred to as the R-4B-R mechanism. These two examples suggest that we can represent mechanisms of various topologies (and sizes) if the stiffness values of the interconnecting zero-length springs are properly adjusted. Therefore, our autonomous synthesis method based on an optimization formulation can be rooted on this unified S-SBM. (Although not explicitly explained here, the stiffness should have its maximum and minimum values at the end of synthesis; otherwise, no distinctly realizable mechanism can be identified, and thus our synthesis formulation will take this aspect into account.) More details are given in Section S1.

**Topology and shape search**

In previous section, we explained that mechanisms of various topologies can be represented by the S-SBM. On the other hand, the shape of a mechanism can be also varied if the joint locations are changed. In the S-SBM, the mechanism shape can be altered by varying the coordinates of the domain-discretizing nodes. As indicated in the Introduction, our autonomous synthesis will be converted as an optimization problem which will be solved by a gradient-based optimizer. To change the topology and shape of a mechanism to be synthesized, we introduce two types of design variables as given below.

*1) Topology-varying design variables:*
$$\xi_m^K \ (m=1,2,\cdots,N_S)$$
($\xi_{min}^K \leq \xi_m^K \leq 1$, $N_s$: total number of zero-length springs in the S-SBM)

The $m^{th}$ zero-length spring stiffness $k_m$ is set to vary as a function of $\xi_m^K$:
$$k_m = k_{max}\left(\xi_m^K\right)^P, \qquad (1)$$

where $p$ is a penalization constant, where $p$ is a penalization constant, $\xi_{min}^K$ is a positive small value, and $k_{max}$ is the maximum stiffness value. Therefore, $\xi_m^K$ is used to determine the topology of a synthesized mechanism.

*2) Shape-varying design variables:*

---

[1] The initial length of the zero-length spring is portrayed as being finite for illustrative purposes only.

$\xi^{\bar{\Theta}}_{2l-1}$ and $\xi^{\bar{\Theta}}_{2l}$ ($l = 1, \cdots, (N_N - 1)$),

($0 \le \xi^{\bar{\Theta}}_{2l-1}, \xi^{\bar{\Theta}}_{2l} \le 1$, $N_N$: total number of nodes in the S-SBM)

The variations $\bar{\Theta}_l$ and $\bar{\Phi}_l$ of the spherical coordinates $\Theta_l$ (polar angle) and $\Phi_l$ (azimuthal angle) used to define the $l^{\text{th}}$ domain-discretizing node in the initial state are varied as the function of

$$\bar{\Theta}_l = \Theta_{\max} \xi^{\bar{\Theta}}_{2l-1} + \Theta_{\min} \left(1 - \xi^{\bar{\Theta}}_{2l-1}\right), \tag{2}$$

$$\bar{\Phi}_l = \Phi_{\max} \xi^{\bar{\Theta}}_{2l} + \Phi_{\min} \left(1 - \xi^{\bar{\Theta}}_{2l}\right), \tag{3}$$

where $\Theta_{\max}$ and $\Theta_{\min}$ ($\Phi_{\max}$ and $\Phi_{\min}$) denote the maximum and minimum variations of the polar angle (azimuthal angle), respectively. Therefore, the location of the $l^{\text{th}}$ node during the synthesis can be expressed as

$$\mathbf{s}_l \left(\Theta_l + \bar{\Theta}_l, \Phi_l + \bar{\Phi}_l\right) = \begin{bmatrix} \sin\left(\Theta_l + \bar{\Theta}_l\right)\cos\left(\Phi_l + \bar{\Phi}_l\right) \\ \sin\left(\Theta_l + \bar{\Theta}_l\right)\sin\left(\Phi_l + \bar{\Phi}_l\right) \\ \cos\left(\Theta_l + \bar{\Theta}_l\right) \end{bmatrix} \quad (l = 1, 2, \cdots, (N_N - 1)). \tag{4}$$

Here, the position of the node to which the input actuator is attached, which is the lower-left corner node of Block 1 in Fig. 2A, is set not be varied during the optimization-based synthesis process. Accordingly, $l$ in Eqs. 2-4 varies from 1 to $(N_N - 1)$. To facilitate subsequent discussions, the following symbol $\xi$ will be used:

$$\xi^T = \left\{\left(\xi^K\right)^T, \left(\xi^{\bar{\Theta}}\right)^T\right\}^T = \left\{\xi^K_1, \xi^K_2, \cdots, \xi^K_{N_S}, \xi^{\bar{\Theta}}_1, \xi^{\bar{\Theta}}_2, \cdots, \xi^{\bar{\Theta}}_{2(N_N-1)}\right\}. \tag{5}$$

**Formulation using the reverse work-transmittance efficiency concept**

Using the design variables introduced above, we can now convert the autonomous synthesis problem into a mathematical optimization problem that can be solved efficiently by a gradient-based optimizer. Specifically, we propose to set up the autonomous synthesis of hip-exo robot mechanisms as the following optimization problem involving the objective function, $\bar{\zeta}(\xi)$ and $N_{const}$ constraint equations, $(\psi^{(i)}_{t^*} - \varepsilon^{(i)} \le 0 \, (i = 1, \cdots, N_{const})$ where $\varepsilon^{(i)}$ denotes a tolerance error and $t^*$, a specific time step):

$$\text{find arg max } \bar{\zeta}(\xi) = \frac{1}{T} \sum_{t^*=1}^{T} \zeta_{t^*}(\xi), \tag{6}$$

$$\text{subject to } \psi^{(i)}_{t^*} - \varepsilon^{(i)} \le 0 \, (t^* = 1, 2, \cdots, T; \, i = 1, 2, \cdots, N_{cnst}), \tag{7}$$

where $T$ is the total time step needed to complete the actuation. The optimization setup expressed by Eqs. 6 and 7 is a standard setup as used in other synthesis problems dealing with fully-actuated mechanisms where their end-effectors are supposed to trace the desired target paths (17-24). However, the objective function employed in previous works (17-24) can be no longer useful in the current synthesis because we need to synthesize under-actuated mechanisms whose end-effectors generate output moments in the required directions. As there is no autonomous synthesis approach suitable for under-actuated mechanisms, regardless of whether an optimization formulation is employed or not, a novel optimization formulation must be developed. This necessitates the development of new objective and constraint functions tailored to the problem posed by underactuated mechanisms.

To establish a new objective function suitable for the optimization-based synthesis of under-actuated mechanisms, we first examine the formulation used in earlier studies (17-24) as it can give us some insight. In these studies, the time-average of the work transmittance efficiency function, $\bar{\eta}(\xi) = \left(\sum_{t^*=1}^{T} \eta_{t^*}(\xi)\right)/T$, was used as the objective function for the synthesis of fully-actuated mechanisms depicted in Fig. 4A. The work transmittance efficiency function $\eta_{t^*}$ is defined as

$$\eta_{t^*} = \frac{W_{t^*}^{out,EF}}{W_{t^*}^{inp,act}} = \frac{W_{t^*}^{out,EF}}{W_{t^*}^{out,EF} + U_{t^*}}, \tag{8}$$

where $W_{t^*}^{out,EF}$ is the output work done by a resistance force applied to the end-effector, $W_{t^*}^{inp,act}$ is the input work provided by the actuators where displacements are given to define the system configuration, and $U_{t^*}$ is the strain energy stored in the elastic elements used to build the unified ground model. It was shown in (17-24) that if $\bar{\eta}(\xi)$ is so maximized as to reach its maximum value of unity, the synthesized mechanism will have a DOF equal to the number of actuators, making it a fully-actuated mechanism[2]. This result suggests that the maximization of $\bar{\eta}(\xi)$ ensures that the mechanism acquires the desired degree of freedom. However, the $\bar{\eta}$ – based formulation cannot be directly used here because the exoskeleton mechanism is an under-actuated system. In an underactuated system, the system DOF is larger than the number of actuators. Therefore, its configuration cannot be defined only with the displacement inputs given to the actuators. Note that previous researches on computational mechanism synthesis considered 1-DOF mechanisms only, but we should deal with mechanisms 3-DOF *underactuated* mechanisms. The differences between fully-actuated and underactuated 3-DOF mechanisms are compared in Fig. 4A and Figure 4B. Figure 4B shows that assistive exoskeleton robots usually adopt an underactuated mechanism since the biological hip joint is a fully actuated system driven by hip muscles on its own and assistive exoskeleton robots only assist part of it. Therefore, even though the idea of the efficiency function employed for the synthesis of fully-actuated mechanisms may be used, an entirely different formulation is required for the underactuated mechanisms being studied in this study.

For the autonomous synthesis of underactuated exoskeleton mechanisms having three degrees of freedom, we newly propose $\bar{\zeta}(\xi) = \left(\sum_{t^*=1}^{T} \zeta_{t^*}(\xi)\right)/T$ as the objective function, where $\zeta_{t^*}$ is defined as

$$\zeta_{t^*} = \frac{W_{t^*}^{out,act}}{W_{t^*}^{inp,EF}} = \frac{W_{t^*}^{out,act}}{W_{t^*}^{out,act} + U_{t^*}}, \tag{9}$$

The core distinction between Eqs. 8 and 9 is that the input and output works are defined in reverse; the input work $W_{t^*}^{inp,EF}$ is defined at the end-effector and the output work $W_{t^*}^{out,act}$ is defined at the actuator in the synthesis problem of under-actuated mechanisms while the input work $W_{t^*}^{inp,act}$ and output work $W_{t^*}^{out,EF}$ are defined at the actuators and the end-effector, respectively, in the fully-actuated mechanism synthesis. To distinguish $\zeta_{t^*}(\xi)$ from $\eta_{t^*}(\xi)$, we will call $\zeta_{t^*}$ the *reverse work transmittance efficiency function*. Note that in this formulation, the mechanism to be synthesized is driven by the generalized displacement at its end-effector, not by the generalized displacements at an input actuator(s). Therefore, the mechanism configuration can be now uniquely defined even if the synthesized mechanism is underactuated; the degree-redundancy issue associated with under-actuated mechanisms cannot be resolved if $\eta_{t^*}(\xi)$ were used.

---

[2] While the employed unified model using elastically-deforming elements allows the representation of mechanisms of different topologies, we need a criterion to tell if a mechanism having a certain distribution of elastic elements becomes a desired mechanism, a one-degree-freedom mechanism in the fully-actuated systems. For its detailed account, please refer to (24, 32, 38).

In order to explicitly define $\zeta_{t^*}$, the generalized rotational displacements $\mathbf{q}_{A,t^*}$ ($t^*=1,2,\cdots,T$) belonging to the orientation workspace of the end-effector are assigned to the end-effector, and an elaborately-defined resistive external generalized force[3] $\mathbf{F}_{t^*}^{ext}$ is applied to the input actuator; see the illustration of Fig. 4C and more details are given in section S2. With this actuation strategy applied to its end-effector, the maximization of $\bar{\zeta}(\xi)$ will ensure that a underactuated 3-DOF mechanism can be obtained, as was shown for the fully-actuated mechanism synthesis (17) where $\bar{\eta}(\xi)$ is maximized. When $\bar{\zeta}(\xi)$ becomes unity, its maximum value, in the present case, $U_{t^*}$, the elastic energy stored in the zero-length springs will become zero.

The output work at the actuator $W_{t^*}^{out,act}$ can be explicitly written as

$$W_{t^*}^{out,act} = \sum_{t=1}^{t^*} \mathbf{F}_t^{ext} \left( \mathbf{q}_{I,t-1} - \mathbf{q}_{I,t} \right), \tag{10}$$

and

$$\mathbf{F}_t^{ext} = \mathbf{T}\mathbf{M}_t^{ext}, \tag{11}$$

where $\mathbf{T}$ is the coordinate transformation matrix and $q_{I,t} = [\rho,\theta,\phi]^T_{\text{at Block } I \text{ at time } t}$ ($\rho$, $\theta$, $\phi$: the Tait-Bryan angles) is the state vector of Block $I$ (the block to which the input actuator is attached). The elastic energy $U_{t^*}$ can be easily calculated by summing up the strain energy stored in all of the zero-length springs.

Next, we will consider the constraint equations. The primary role of the constraint equations is to ensure that the synthesized mechanism generates the target moment of the desired varying directions at every time step of its wearer. Specifically, the output moment $\mathbf{M}_{t^*}^{out,EF}$ generated by the end-effector of the mechanism being synthesized should be the same as the target output moment $\hat{\mathbf{M}}_{t^*}^{out,EF}$ for a given input moment $\mathbf{M}_{t^*}^{inp,act}$ at time $t^*$ ($t^*=1,2,\cdots,T$). To indicate that the direction of the input moment at the actuator remains fixed, we express $\mathbf{M}_{t^*}^{inp,act}$ as $\mathbf{M}_{t^*}^{inp,act} = \left| \mathbf{M}_{t^*}^{inp,act} \right| \mathbf{g}^{inp,act}$ where $\mathbf{g}^{inp,act}$ is the direction-fixed unit vector representing the input moment direction. To measure the difference in the direction and magnitude between the output moment $\mathbf{M}_{t^*}^{out,EF}$ and the target output moment $\hat{\mathbf{M}}_{t^*}^{out,EF}$, we may consider use two constraint functions $\psi_{t^*}^{(1)}$ and $\psi_{t^*}^{(2)}$ given by

$$\text{direction:}\quad \psi_{t^*}^{(1)} = \left( \left( \frac{\mathbf{M}_{t^*}^{out,EF}}{\left|\mathbf{M}_{t^*}^{out,EF}\right|} \cdot \frac{\hat{\mathbf{M}}_{t^*}^{out,EF}}{\left|\hat{\mathbf{M}}_{t^*}^{out,EF}\right|} \right)^2 - 1 \right), \tag{12a}$$

---

[3] This generalized force does not represent the moment needed to drive the actuator; it is a static (dead) moment introduced to define $\zeta_{t^*}$.

$$\text{magnitude:} \quad \psi_{t*}^{(2)} = \left\| \mathbf{M}_{t*}^{out,EF} \right| - \left| \hat{\mathbf{M}}_{t*}^{out,EF} \right\| / \left| \mathbf{M}_{t*}^{inp,act} \right|, \tag{12b}$$

These functions appear to be the most intuitive selections, but the direct use of the moment-transmittance condition may hinder the synthesis of a mechanism having the correct 3 DOF, compared with the use of displacement conditions used earlier in synthesizing mechanisms generating the desired path (or desired kinematic motions) (17-24). The reason is that constraint equation 7 can be satisfied even when the DOF of the synthesized mechanism can be larger than $3^4$; when DOF redundancy occurs, the $\bar{\zeta}$ value can be maximized to a value smaller than unity, and thus an underactuated 3-DOF mechanism cannot be obtained even though these constraint equations are satisfied. Therefore, we need an alternative formulation to implicitly push the $\bar{\zeta}$ value towards unity as the optimization for mechanism synthesis converges.

Inspired by the fact that the successful displacement-based formulation (see (17-24)) implicitly suppresses DOF redundancy, we propose the following strategies. First, we convert the moment-transmittance condition between the input link (or input block) and the end effector into an angular velocity-transmittance condition. For the conversion, the principle of virtual work is used:

$$\left( \mathbf{M}_{t*}^{out,EF} \right)^T \delta \mathbf{\Theta}_{t*}^{EF} = \left( \mathbf{M}_{t*}^{inp,act} \right)^T \delta \mathbf{\Theta}_{t*}^{act}, \tag{13}$$

where

$$\delta \mathbf{\Theta} = [\delta \theta_X, \delta \theta_Y, \delta \theta_Z]^T \ (\delta \theta_\alpha : \text{virtual rotation about the } _\alpha\text{-axis}).$$

Using Eq. 13, $\delta \mathbf{\Theta}_{t*}^{EF}$ and $\delta \mathbf{\Theta}_{t*}^{act}$ will be used as the field variables as alternatives to $\mathbf{M}_{t*}^{out,EF}$ and $\mathbf{M}_{t*}^{inp,act}$. Then, the perturbed angular displacement $\delta \mathbf{\Theta}_{t*}^{act}$ of the actuator (or block *I* to which it is attached) should be calculated subject to the perturbed angular displacement $\delta \mathbf{\Theta}_{t*}^{EF}$ of the end-effector at every time step $t*$. The end-effector is perturbed from its current pose by $\delta \mathbf{\Theta}_{t*}^{EF}$ with respect to the global $X$, $Y$, and $Z$ axes shown on the Figure 4D. Since a resistive moment ($\mathbf{M}_{t*}^{ext}$) is being applied to the block *I* for the calculation of $\bar{\zeta}$ and the input actuator is attached to the ground model via a zero-length elastic spring, the direct measurement of the angular displacement change of the end-effector may be insufficiently accurate. To avoid this difficulty, we propose to calculate $\delta \mathbf{\Theta}_{t*}^{act}$ using the perturbed translation displacements of three points ( $\mathbf{p}_{1,t*} = [1,0,0]^T$, $\mathbf{p}_{2,t*} = [0,1,0]^T$, and $\mathbf{p}_{3,t*} = [0,0,1]^T$ ) on the spherical surface extended from Block *I*. If the perturbed displacement components $\left[ \Delta(\mathbf{p}_{j,t*})_X, \Delta(\mathbf{p}_{j,t*})_Y, \Delta(\mathbf{p}_{j,t*})_Z \right]_{j=1,2,3}$ are found for the prescribed $\delta \mathbf{\Theta}_{t*}^{EF}$ at the end-effector, they can be compared with their target values $\left[ \Delta(\mathbf{p}_{j,t*})_X, \Delta(\mathbf{p}_{j,t*})_Y, \Delta(\mathbf{p}_{j,t*})_Z \right]_{j=1,2,3}$. Accordingly, we propose to use the following constraint functions that are equivalent to those in Eq. 12:

---

[4] Although the degree of freedom should be an integer in actual mechanisms, the degree of freedom of the mechanism represented by the S-SBM, a ground model in use, may be regarded as being somewhere between two consecutive integer values because elastic elements are used in the ground model.

$$\begin{cases} \psi_{t^*}^{(j)} = \|\Delta(\mathbf{p}_{j,t^*})_X - \Delta(\hat{\mathbf{p}}_{j,t^*})_X\| \\ \psi_{t^*}^{(j+3)} = \|\Delta(\mathbf{p}_{j,t^*})_Y - \Delta(\hat{\mathbf{p}}_{j,t^*})_Y\| \quad \text{for } j = 1, 2, 3. \\ \psi_{t^*}^{(j+6)} = \|\Delta(\mathbf{p}_{j,t^*})_Z - \Delta(\hat{\mathbf{p}}_{j,t^*})_Z\| \end{cases} \quad (14)$$

In actual numerical calculations, the angle variation of $\delta\Theta_{t^*}^{EF}$ along each of the three global coordinate axes ($X,Y,Z$) is set to $10^{-3}$, which is sufficiently small. For more detailed accounts of the formulation, refer to Section S3. Section S4 describes the sensitivity of the objective and constraint functions necessary for the application of a gradient-based optimizer.

**Description of synthesis problems**

We will now tackle three synthesis problems involving the synthesis of single-actuator-driven hip-exo robot mechanisms that satisfy diverse design constraints, such as varied output moment profiles and ranges of motion, using our autonomous synthesis method. In the main part of this paper, we will present the synthesis for Case Study 1, which deals with the output moment profile shown in Fig. 2B. Before presenting the result for Case Study 1, it is remarked that the results for two other cases, Case studies 2 and 3, dealing with the different moment profiles and/or ranges-of-motion, are given in Tables S1 and S2, respectively, in Supplementary Materials. The synthesis of a well-known mechanism that produces the output moment of a fixed direction during all gait phases—in this example, the abduction direction—is the focus of Case Study 2, in particular. In this case, the target output moment is $\hat{\mathbf{M}}_{t^*}^{out,EF} = M_0[0,-1,0]^T$ where $M_0$ is the output moment magnitude that is equal to the magnitude of the input torque from the actuator. The synthesis process is presented in Fig. S2. Our autonomous synthesis method successfully recovered the well-known series mechanism (which will be called the R-R-R mechanism) shown at the bottom of Fig. S2A where three revolute joints (R's) are connected in series. Series mechanisms have been widely used in existing gait-assistance exoskeleton robots (2, 26, 27, 32, 34). We will skip the detailed accounts of the results in Fig. S2[5], but this example shows that a widely-used serial mechanism can also be synthesized by our autonomous synthesis method if the user wants. As Case Studies 1 and 3 deal with the direction-varying output moment profiles, we anticipate to obtain mechanisms that are vastly dissimilar to these R-R-R mechanisms. Here, we will mainly discuss the synthesis of Case Study 1 in some detail while the results for Case Study 3 are given in Fig. S3. Comparison of the results for Case Studies 1 and 3 show that depending on the specified individual's requirments, mechanisms of different topologies can be found by ExoSynthesizer when a direction-varying output moment is desired.

To deal with Case Study 1 seeking to create a single-actuator-driven mechanism that generates the output profile given in Fig. 2B, we use the data in Fig. 2. They are selected as representative values from the literature (42). Accordingly, it is assumed (42) that $\phi_{A,t^*} = 0.4$ rad ($\approx -20°$) at $t^* = 5$, $\phi_{A,t^*} = -0.4$ rad ($\approx 20°$) at $t^* = 1$, and $\phi_{A,t^*} = 0$ rad at $t^* = 3$, respectively. Here, the subscript $A$ refers to the block to which the end-effector is attached. To assist hyper-flexed poses possibly occurring for other gait modes such as incline walking, we set $\phi_{A,t^*} = 0.8$ rad at $t^* = 7$. At other time steps $t^* = 2, 4,$ and $6$, the $\phi_{A,t^*}$ values are equally distributed, while the rotation and abduction/adduction angles are assumed to be zero for all time steps ($\rho_{A,t^*}, \theta_{A,t^*} = 0$).

---

[5] Once we fully discuss Case study 1 below, the meaning of various illustrations in Figs. S3 and S2 will be clearly understood.

For the synthesis, The input moment is assumed to be given as $\mathbf{M}_{t^*}^{inp,act} = M_0 [0, -1, 0]^T$. The target output moments $\hat{\mathbf{M}}_{t^*}^{out,EF} = M_0 [\kappa_X, -1, \kappa_Z = 0]^T$ are described by $\kappa_X$ listed in Fig. 2B for the selected poses at different time steps. The symbol denotes is the ratio of the flexion/extension component to the abduction component. Note that the rotation component of the target output moment is assumed to be zero. The output abduction/adduction moment component ($-M_0$) is set to have the same magnitude as the input moment because the moment equilibrium condition must be satisfied. To explain how the values of $\kappa_x$ are chosen, we consider the bottom illustration of Fig. 2A, which depicts the direction of the biological hip moment of a human according to his or her gait phase governed by the position of his/her femur. Note that the gait phases of the initial contact and loading responses correspond to the time steps of $t^* = 4$ and $t^* = 5$ while the hyper-flexed poses correspond to the time steps of $t^* = 6$, $t^* = 7$. The hip moment resultants consisting of flexion/extension and abduction moment components are indicated by big arrows in the $X-Y$ plane[6] in Fig. 2A. The extension moment component with respect to the abduction moment component is set to increase from $t^* = 4$ to $t^* = 7$ to reflect the fact that the moment arm for the GRF (ground reaction force) becomes larger in the sagittal plane while it remains nearly constant in the frontal plane as a human flexes his or her leg (see, e.g., Perry and Burnfield (42)). Notably, the hip moment has nearly zero flexion/extension moment at the mid stance phase ($t^* = 3$) because the GRF passes through the center of the hip joint in the sagittal plane. (See the bottom illustration in Fig. 2A.) The terminal and pre-swing phases are in the period between $t^* = 2$ and $t^* = 1$. During these phases, the flexion moment component with respect to the abduction moment component increases as a human extends his or her leg backwards. Note that, to show if a mechanism generating varying-moment having a more significant flexion/extension component, having larger $\kappa_X$ values, could be synthesized from our method, Case Study 3 is dealt with.

Once the topology and initial shape of joint mechanisms for the given output moment profiles are determined using our autonomous method, they may be modified to better suit the requirements of each individual. This will be discussed in the "Biomechanical assessment of the synthesized hip mechanism" section.

**Synthesis process by optimization formulation**
Figure 5A shows the history of the autonomous mechanism synthesis for Case Study 1 and the post-processed synthesized mechanism. (Those for Case Studies 2 and 3 are shown in Figs. S2 and S3, respectively.) The first row of the table in Fig. 5A shows the top view of the motion of the wearer's right leg at time steps =1, 3, 5, and 7. The initial and converged configurations of the spherical-SBM are presented in the 2nd and 5th rows, while two intermediate configurations during the synthesis are shown in the 3rd and 4th rows. The iteration history shows that the output moment generated by the mechanism being synthesized, represented by a blue arrow, gradually approaches the target output moment, represented by a red arrow, as the synthesis converges.

---

[6] The rotation moment is assumed to be zero.

Before delving further into the synthesis history for Case Study 1, we remark that the synthesized mechanism is entirely different from the R-R-R mechanism creating a direction-fixed output moment, demonstrating why the ability to explore various topologies in our methodology is so important. The synthesized mechanism by ExoSynthesizer is a novel mechanism that serially consists of a revolute joint (R), a closed 4-bar mechanism (will be denoted by 4B), and another revolute joint (R); no mechanism of similar configuration has been used in exoskeleton robots. The resulting mechanism, which will be denoted by the R-4B-R mechanism, is a 3-DOF hybrid serial-parallel mechanism that replaces the 2nd revolute joint, which is not connected to the upper harness nor to the thigh harness, by a 4-bar mechanism. It is stressed that the 4-bar mechanism formed between two revolute joints is essential in producing the direction-varying output moment, even though we will go into more detail about how the R-4B-R mechanism can generate the direction-varying moment in the section titled "Kinematic analysis of the synthesized 'R-4B-R' mechanism." As Case study 3 considers a different moment-direction profile with a larger flexion/extension moment than Case study 1 for a slightly modified pose, our ExoSynthesizer generates another novel mechanism consisting of a revolute joint (R), a four-bar mechanism (4B), and a five-bar mechanism (5B). It will be denoted by the R-4B$\oplus$5B mechanism. Here, we introduce a symbol "$\oplus$" in naming the resulting mechanism because the four- and five-bar mechanisms are uniquely intertwined in parallel. It is shown in Fig. S3A. Despite the R-4B$\oplus$5B mechanism may seem complex for practical applications, this result demonstrates that our method may provide a solution even when the user requests design requirements that may be difficult to meet. The configurations of the synthesized mechanisms for the right hip joint can be easily mirrored for the left hip joint.

To check the convergence of the ExoSynthesizer algorithm, we first examine the evolution of the reverse work transmittance efficiency values $\bar{\zeta}$ denoted by a red line in Fig. 5B. It gradually increases from nearly zero at the initial iteration stage to nearly 1, its theoretical maximum value, at the converged state. This confirms that the synthesized hip-exo robot mechanism has the desired 3 degrees of freedom at convergence. Next, we examine the errors in the output moment direction using $\max\left(\psi_{t*}^{(1)}, \psi_{t*}^{(2)}, \psi_{t*}^{(3)}\right)$, $\max\left(\psi_{t*}^{(4)}, \psi_{t*}^{(5)}, \psi_{t*}^{(6)}\right)$, and $\max\left(\psi_{t*}^{(7)}, \psi_{t*}^{(8)}, \psi_{t*}^{(9)}\right)$, which represent how accurately the desired flexion/extension, abduction/adduction, and rotation moment components are traced, respectively. They are plotted in Figure 5B by a black solid line, a black dashed line, and a blue solid line, respectively. (The same notations are used in Figs. S2B and S3B for Case Studies 2 and 3, respectively.) Relatively large errors occurred in $\max\left(\psi_{t*}^{(1)}, \psi_{t*}^{(2)}, \psi_{t*}^{(3)}\right)$ and $\max\left(\psi_{t*}^{(7)}, \psi_{t*}^{(8)}, \psi_{t*}^{(9)}\right)$ at initial iterations because the mechanism at the initial configurations produces the corresponding moment components quite different from their target values. On the other hand, nearly zero error in $\max\left(\psi_{t*}^{(4)}, \psi_{t*}^{(5)}, \psi_{t*}^{(6)}\right)$ occurred because the formation of additional ground revolute joints other than the one used to connect the input actuator to the ground (the human body) was prohibited in our modeling. When the synthesis process was completed, all errors in these functions were found to fall within the prescribed tolerance error, $\varepsilon^{(i)} = 2 \cdot 10^{-4}$, which corresponds to a difference of approximately 10 degrees between the target and actual moment directions. The use of smaller tolerance error would result in mechanisms satisfying the conditions on the output moment direction better, but it may be difficult to obtain converged results. If necessary, users could additionally modify shape modification of the synthesized mechanisms can be performed: The convergence pattern is generally stable and fast except for some iteration numbers around 40~80 and around 110~120. The unstable behavior, however, really demonstrates how well the displacement-based formulation suppresses DOF redundancy implicitly, as is covered in the section headed "Formulation using the reverse work-transmittance efficiency concept." The employed optimization solver, called the method of moving asymptotes (MMA) (43), prioritizes satisfying the constraint equations before attempting to maximize or minimize the objective function. The synthesized mechanism undergoes a wide range of topological alterations when the objective function is optimized. The constraint function abruptly rises above the tolerance limit if the system becomes redundant in terms of degrees of freedom (DOF > 3) during the topology modifications because it can no longer withstand external resistive moments. The optimizer adjusts the degrees of freedom to lower the constraint function in order to handle this. If the direct moment error measurement in Eqs. 12a and 12b was used as constraint functions, the system would not converge to the required DOF, as it cannot capture the DOF redundancy.

Next, we will look into the spring stiffness values. Although not explicitly shown here, our ExoSynthesizer algorithm gradually adjusted the stiffness values of the zero-length springs that determine the topology of the mechanisms, which in turn determine the presence and location of revolute joints. The algorithm also varied the positions of the block nodes, which in turn determine the positions (or the directions of rotation axes) of the revolute joints. Figure 5A shows how the shapes of blocks are continuously varied. As for the stiffness values of zero-length springs, which were initially set to intermediate values $k_m = k_{max}\left(\xi_m^k\right)^p = k_{max}(1/2)^p$, they approached sufficiently close to their maximum ($k_{max}$) or minimum (~0) values. Therefore, the synthesized mechanism found at the converged state can be identified without ambiguity.

**Kinematic analysis of the synthesized 'R-4B-R' mechanism**
We will now explain how the synthesized R-4B-R mechanism can transmit a fixed-direction input moment to a direction-varying output moment in response to an input gait motion. Prior to conducting its kinematic analysis, we examine its topological connectivity. Referring to Fig. 6A, we denote the R joint connected to the upper harness by R1. It shares the same axis as the input rotary actuator. R2 and R4 denote two other revolute joints belonging to the link to which R1 is attached. Thus, R1, R2, and R4 are connected to the link, called L1. The revolute joint R2 is connected to R3 through the link L2, and the revolute joint R4 is connected to R5 through the link L3. On the other hand, R3 and R5 are connected to another link L4 while another revolute joint R6 is connected to it. The thigh harness is connected to the link L4 through R6. This connectivity analysis shows that the links L1, L2, L4, and L3 along with the revolute joints R2, R3, R5, and R4 form a four-bar mechanism. Therefore, the synthesized mechanism can be interpreted as a serial connection of a revolute joint, a four-bar mechanism, and another revolute joint.

Now, we will analyze the direction of the output moment generated by the synthesized R-4B-R mechanism by the screw-axis theory's actuation wrench (44). For the analysis, we view the 3-DOF R-4B-R mechanism driven by a single actuator as one of three limbs of a fully actuated 3-DOF parallel mechanism. For this analysis, the twists for the revolute joints R1, R2, R3, R4, R5, and R6 are defined as $\$_1$, $\$_2$, $\$_3$, $\$_4$, $\$_5$, and $\$_6$, respectively. Because each revolute joint allows pure rotational motion only with respect to its axis penetrating the origin of the spherical coordinate system, $\$_i$ can be expressed as

$$\$_i = \left[\mathbf{r}_i^T; 0,0,0\right]^T \ (\text{for } i=1,2,\cdots,6) \tag{15}$$

where $\mathbf{r}_i$ denotes a unit 3×1 rotation axis vector of the revolute joint R*i*. The link L4 allows a 1-DOF instantaneous pure rotation motion with respect to the link L1 because the four revolute joints, R2-R5, form a 1-DOF four-bar mechanism. The corresponding instantaneous rotation axis $\mathbf{r}_v$ can be found as

$$\mathbf{r}_v = \frac{(\mathbf{r}_2 \times \mathbf{r}_3) \times (\mathbf{r}_4 \times \mathbf{r}_5)}{\|(\mathbf{r}_2 \times \mathbf{r}_3) \times (\mathbf{r}_4 \times \mathbf{r}_5)\|}, \tag{16}$$

which is the intersection of two great circles spanned by [$\mathbf{r}_2, \mathbf{r}_3$] and [$\mathbf{r}_4, \mathbf{r}_5$]. Consequently, the twist of the instantaneous rotation motion of link L4 with respect to link L1 can be defined as

$$\$_v = \left[\mathbf{r}_v^T; 0,0,0\right]^T, \tag{17}$$

the direction of which is illustrated in Figure 6A. Accordingly, the R-4B-R mechanism can be viewed as being instantaneously identical to a 3-DOF serial mechanism consisting of R1, a virtual revolute joint VR, and R2, as indicated in Figure 6B. Here, the virtual revolute joint VR is defined as a revolute joint whose axis is the same as the instantaneous rotational axis $\mathbf{r}_v$ of the 4-bar mechanism formed by the four revolute joints R2-R5.

Here, the actuation moment, or more generally, the actuation wrench $\mathcal{W}_a$, which is a moment generated from the end effector of this mechanism to the wearer when the input actuator is actuated, can be calculated by the following relation (44):

$$\mathcal{W}_a = \mathcal{W}_\varsigma - \mathcal{W}_c \tag{18}$$

where $\mathcal{W}_\varsigma$ is a wrench system spanned by the wrench screws that the end effector can resist after locking the input actuator, and $\mathcal{W}_c$ is a constraint wrench system spanned by the wrench screws that the end effector can resist without locking any joints. Note that $\mathcal{W}_\varsigma$ and $\mathcal{W}_c$ are reciprocal to twist systems $\mathcal{T}_\varsigma = \text{span}(\$_6, \$_v)$ and $\mathcal{T}_c = \text{span}(\$_1, \$_6, \$_v)$, respectively. All axes of revolute joints or a virtual revolute joint pass through the origin $O=[0,0,0]$ and their axes vectors $\mathbf{r}_1$, $\mathbf{r}_6$, and $\mathbf{r}_v$ are linearly independent because the mechanism configuration is non-singular through all time steps. Therefore, we can derive $\mathcal{W}_\varsigma$ and $\mathcal{W}_c$ as

$$\mathcal{W}_\varsigma = \text{span}\left([1,0,0;0,0,0]^T, [0,1,0;0,0,0]^T, [0,0,1;0,0,0]^T, [0,0,0;\mathbf{r}_6 \times \mathbf{r}_v]^T\right),$$

$$\mathcal{W}_c = \text{span}\left([1,0,0;0,0,0]^T, [0,1,0;0,0,0]^T, [0,0,1;0,0,0]^T\right).$$

The actuation wrench $\mathcal{W}_a$ can be written as $\text{span}\left([0,0,0;\mathbf{r}_6 \times \mathbf{r}_v]^T\right)$, and it denotes a pure moment with the direction of $\mathbf{r}_6 \times \mathbf{r}_v$. This means that the output moment transmitted from the input actuator by the synthesized mechanism is a pure moment in the direction perpendicular to the great circle connecting the revolute joint R6 and the virtual revolute joint VR shown in Fig. 6B.

Figure 6C provides additional insight into the kinematic behavior of the synthesized R-4B-R mechanism. It depicts the snapshots of the movements of the equivalent serial mechanism sketched on the right side of Fig. 6B and the link lengths of the virtual links VL1 and VL2 during the gait phases of the wearer. As shown in Fig. 6C, the location of the virtual link VL2 (marked in blue) varies over time steps, but it is fairly vertically aligned. Because the axis of the mechanism's output moment is perpendicular to the great circle containing VL2, the axis direction changes with a nearly zero rotation moment component. Moreover, the link lengths of VL1 and VL2 vary according to the time steps, as shown in the extreme right side of Fig. 6C. Note that the change in the lengths of VL1 and VL2 cannot be possible if no virtual revolute joint VR can be envisioned. The top view of the left illustration of Fig. 6C is depicted on the left side of Fig. 6D. These analyses support that the desired directions of the output moment are indeed obtained by the R-4B-R mechanism; the output moment has the combined extension and abduction direction at the flexed leg poses ($t^*=4\sim7$), the abduction direction (with a small amount of flexion component within tolerated error) at the neutral leg pose ($t^*=3$), and the combined flexion and abduction direction at the extended leg poses ($t^*=1\sim2$). The latitude and longitude of the axis of the output moment generated by the synthesized mechanism at different gait phases (or time steps) are plotted on the right side of Fig. 6D, indicating that the latitude and longitude obtained by the synthesized mechanism are nearly identical to their target values. All of these analyses confirm that the direction-varying output moment for a direction-fixed input moment cannot be generated without the R-4B-R mechanism; this output moment cannot be produced by conventional hip mechanisms because their link lengths are always fixed. In contrast, the R-4B-R mechanism functions as if it has two length-varying virtual links so that it can change the output moment's direction based on the wearer's leg poses.

**Biomechanical assessment of the synthesized hip mechanism**
The musculoskeletal analysis software OpenSim (41) is used to investigate the assistive capability of a wearable hip robot equipped with the R-4B-R mechanism. The OpenSim gait2392 model and a sample gait motion were utilized for the analysis. For the investigation, the shape of the synthesized mechanism has been modified to accommodate a particular target individual whose gait pattern requires a small rotational output moment component and a large range of extension motion. The revolute joint axes before and after modification are listed in Table S3, and the sample gait motion is discussed in Section S5. The mechanism for the left hip joint was created by mirroring the right mechanism with respect to the median plane of the mechanism wearer.

Figure 7A displays the output moment profiles that the synthesized mechanisms with the aforementioned modifications will produce. The left and right graphs correspond to the mechanisms [7] used for the right and left hips, respectively. Once the output abduction/adduction moment components that are the same as the input torque profiles from the input actuators are produced, as depicted in Fig. 7A in red, the R-4B-R mechanisms generate the remaining moment components. The rationale for our choice of the input torque profiles is given in Section S5. The output moment profile for the right mechanism resembles the biological hip moment profile sketched in Figure S1 in that the abduction moment profile in red has double peaks, the flexion/extension direction moment in black is along the extension direction before the mid-stance phase while it is along the flexion direction after the mid-stance phase, and the rotation direction moment is nearly zero throughout the entire gait cycle. The moment profile of the left hip joint resembles the biological hip moment profile, just like the right hip joint, but it seems different because the left and right joints have different gait phases.

---

[7] They will be simply referred to as the right and left mechanisms.

The effects of gait assistance provided by the R-4B-R mechanisms are evaluated using the CMC (computed muscle control) module in OpenSim. The activation of all lower-body muscles required to generate the predefined gait motion was calculated, and the analysis results are presented in Fig. 7B. The figure shows that both the peak extension/flexion and abduction moments are simultaneously reduced, which is not possible with a direction-fixed single input actuation. The peak value in the abduction moment is most significantly reduced as programmed. Fig. 7B also demonstrates that the assisted and unassisted biological moments in the rotational direction are nearly identical, indicating that the wearer is not adversely affected in the rotation direction by the use of the R-4B-R mechanism. The details of the fabricated prototype of the synthesized R-4B-R mechanism, shown in the prototype part of Figure 1, are given in Section S6.

## DISCUSSION

Here, we developed an autonomous computational method using gradient-based optimization for wearable robot mechanisms. The autonomous method, the first of its kind that does not require the use of a specific baseline mechanism, simultaneously determines the topology and shape of desired mechanisms satisfying the specified design requirements. To maintain the multi-DOF underactuated nature of exoskeleton mechanisms, a new formulation based on the reverse work transmittance efficiency is proposed. The potential of the proposed method is demonstrated using several hip-exo robot design problems, including the problem of designing novel hip-exo robot mechanisms that generate direction-varying output moments according to the gait phase of a human under a single direction-fixed input actuation. The autonomous method successfully synthesized mechanisms with different hybrid serial-parallel connectivity, such as R-4B-R, depending on different output moment profiles and ranges of motion. All the results were obtained in around tens of minutes on a desktop computer. Based on the musculoskeletal analysis, the R-4B-R mechanism was found to effectively assist hip moments by reducing the peak values of the wearer's biological hip moments in both the sagittal and frontal planes. The basic wearability of the device was also evaluated via prototyping. As this study's scope is limited to the formulation and computational aspects of autonomous synthesis of wearable robot mechanisms, human testing of the synthesized mechanisms will be conducted in the future. Although the autonomous synthesis of hip joint mechanisms generating direction-varying moments from a single actuator was the primary focus of this study, the developed method can also be applied to other parts of wearable robots where linkage mechanisms can be used as intelligent force/moment transmitters. For example, a self-aligning mechanism that should transform input force/moment into a pure moment on a joint over a wide range of motion is an interesting application direction. Finally, our method will be able to be a core algorithm for realizing interactive design tools for personalized/customized wearable robot development pipelines thanks to its agile and creative mechanisms searching capability.

## MATERIALS AND METHODS

### Design domain discretization with spherical-SBM

In the part titled "Modeling with spherical-SBM," $\Phi_P = \pi/2$ and $\Theta_T = \pi/4$ were used to define the design domain; refer to Fig. 2A. We excluded the spherical surface region of $\Theta < \pi/4$ and $\Phi > 0$ to prevent the synthesized mechanism from interfering with arms. In addition, we excluded the spherical surface region of $\Theta > \pi/2$ to allow the wearer to sit and the region of $\Phi < -\Phi_P$ to prevent the mechanism from interfering with the left hip joint mechanism.

The design domain is uniformly discretized by $N_B = N_P \times N_T$ two-dimensional spherical rigid blocks with the same latitude and longitudinal size, where $N_P$ and $N_T$ denote the numbers of blocks along the azimuth and altitude directions, respectively. The spherical blocks are interconnected to each other by $N_S$ zero-length block-connecting springs at $N_N$ nodes used to define the blocks. In the present investigation, $N_P = 4$, $N_T = 2$, $N_B = 8$, $N_S = 20$, and $N_N = 15$ were used as depicted in Fig. 2A. Note that the input actuator (providing a input moment) is attached to Block $I$ (=1), which is attached to the upper body harness, and the end-effector is connected to Block $A$ (=4), which is attached to the thigh harness.

**Selected parameter values**
The following values of the parameters appearing in the equations in the main text and supplementary materials were used for the numerical implementation of our algorithm:
$$p = 3,\ \xi_{min}^K = 0.001,\ k_{max} = 10^4,\ F_0 = 1,$$
$$\varepsilon^{(i)} = 2 \cdot 10^{-4},$$
$$\Phi_{max} = \pi/8,\ \Phi_{min} = -\pi/8,\ \Theta_{max} = \pi/8,\ \Theta_{min} = -\pi/8,$$

The parameters $k_{max}$, $F_0$, and $\xi_{min}^k$ are needed to calculate the reverse work transmittance efficiency function, and their values are selected with reference to previous studies (31, 32, 34, 35, 38) that used the work transmittance efficiency function for fully-actuated mechanisms. Other values were selected after numerical tests. The parameter $\varepsilon^{(i)}$ is used to control the error in the output moment direction, and the selected value of $\varepsilon^{(i)} = 2 \cdot 10^{-4}$ corresponds to around $10°$ deviation from the actual and target output moment directions. The parameters $\Phi_{max}$ ($\Phi_{min}$) and $\Theta_{max}$ ($\Theta_{min}$) determine the maximum (minimum) movement of the domain-discretizing nodes in the spherical coordinates. All nodes are allowed to vary by the same amount. To avoid interference of the synthesized mechanism with the wearer, however, all nodes are prohibited not to move outside of the design domain in the altitude direction. In Case study 3, we used slightly smaller values ranging between $0.45 \times (-\pi/8)$ and $0.45 \times (\pi/8)$ (for both $\Phi$ and $\Theta$ values); we chose smaller values in this case because singular SBM configurations tend to appear because Case study 3 allows a larger variation in the output moment direction. To prevent a biased search for the optimal solution, all design variables were initialized to their median values (0.5) prior to solving the optimization problem using a gradient-based optimizer.

**Supplementary Materials**
Tables S1 to S4
Figures S1 to S6
Sections S1 to S6

**References and Notes**
1. M. Attaran, The rise of 3-D printing: The advantages of additive manufacturing over traditional manufacturing, *Business Horizons* **60**, 677-688 (2017).

**Acknowledgments:** We would like to express our gratitude to Dr. Dong Jin Hyun and his group from the Research and Development Division of Hyundai Motor Company for providing the initial inspiration for the design of the gait assistive hip exoskeleton robot.

**Funding:** This work was supported by the Samsung Research Funding & Incubation Center of Samsung Electronics under Project Number. SRFC-IT1901-02

**Author contributions:** Seok Won Kang and Yoon Young Kim conceived the research. Seok Won Kang defined the synthesis problems, implemented the synthesis algorithm, and did a kinematic analysis of the synthesized mechanism. Seok Won Kang and Suh In Kim devised the optimization formulation. Jegyeong Ryu and Seok Won Kang conducted a biomechanical assessment of the synthesized mechanism via simulation. Seok Won Kang and Youngsoo Kim fabricated the prototype of the synthesized R-4B-R mechanism. Yoon Young Kim supervised the overall research. The manuscript with the results was reviewed by all the authors.

**Competing interests:** Patents for the synthesized mechanisms given in this paper have been filed by Seoul National University (Application, KR 2020-0136264, US 17/154,086, CN 202110499359.8; Provisional Application, KR 10-2023-0016956). Other than the patents, there are no conflicts of interest to declare.

**Data and materials availability:** Main data and materials are given in the paper and/or the Supplementary Materials. Additional data are available from the corresponding authors upon reasonable request.

**Figures:**

**Fig. 1. Illustrative flow chart of the 'ExoSynthesizer' and fabricated prototype.** As inputs, ExoSynthesizer receives the desired range of motion (or workspace) specified by discretized points (or poses) and the desired force/moment transmission properties at each discretized point. After rapidly and iteratively updating design variables with a gradient-based algorithm, ExoSynthesizer provides its users with the topology and size of the synthesized mechanisms. New mathematical models and formulations were proposed to perform the autonomous synthesis of underactuated exoskeleton robot mechanisms.

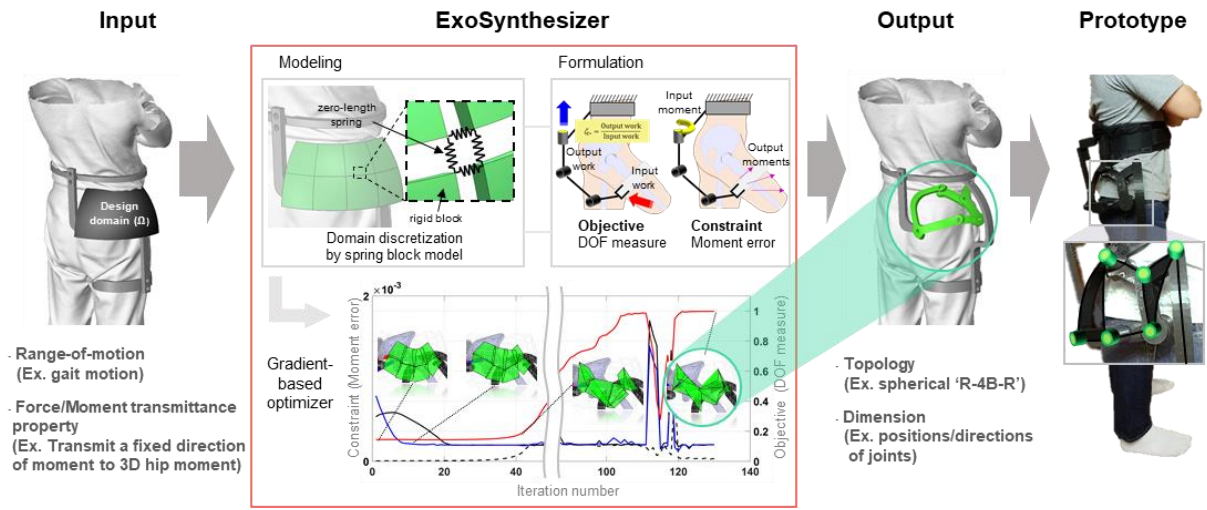

**Fig. 2. Hip-exo robot mechanism synthesis problem (A)** The design domain and the illustration of the variation of the output moment direction, and **(B)** the target range of motion (given by degrees) and output moment profile for Case Study 1.

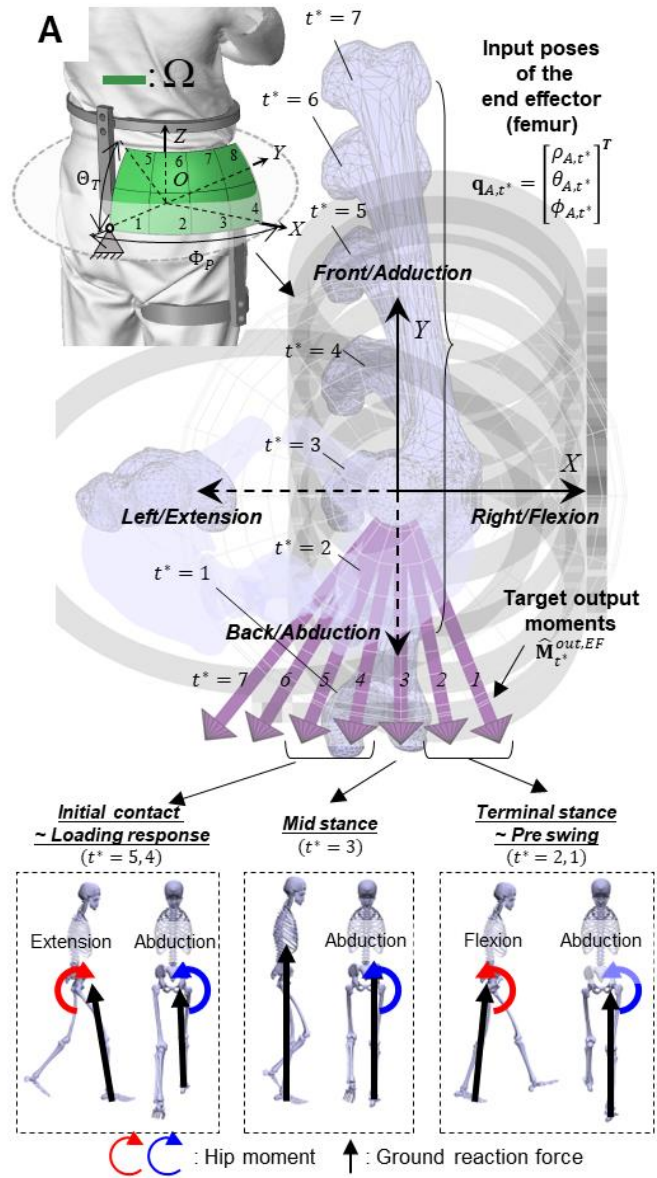

| $t^*$ | $(\rho_{A,t^*}, \theta_{A,t^*}, \varphi_{A,t^*})$ (rad) | $\widehat{\mathbf{M}}_{t^*}^{out,EF}/M_o$ |
|---|---|---|
| 1 | (0, 0, -0.4) | ($\kappa_X$=-0.4, -1, 0) |
| 2 | (0, 0, -0.2) | ($\kappa_X$=-0.2, -1, 0) |
| 3 | (0, 0, 0) | ($\kappa_X$=0, -1, 0) |
| 4 | (0, 0, 0.2) | ($\kappa_X$=0.2, -1, 0) |
| 5 | (0, 0, 0.4) | ($\kappa_X$=0.4, -1, 0) |
| 6 | (0, 0, 0.6) | ($\kappa_X$=0.6, -1, 0) |
| 7 | (0, 0, 0.8) | ($\kappa_X$=0.8, -1, 0) |

**Fig. 3. Newly developed spring-connected spherical rigid block model (Spherical-SBM). (A)** A representation of a revolute joint using zero-length springs and blocks and **(B)** various spherical linkage mechanisms that can be constructed from the proposed spherical-SBM.

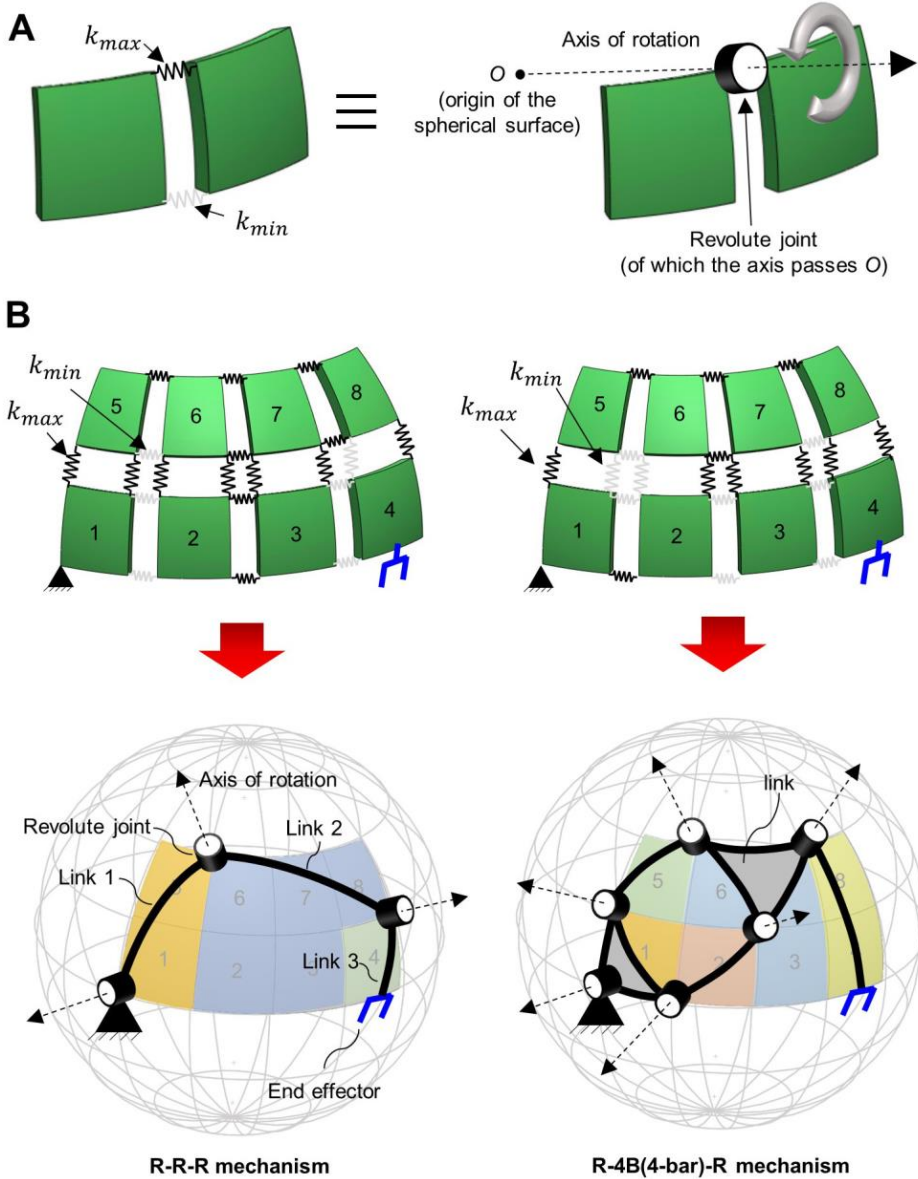

**Fig. 4. DOF-related optimization formulation issue for the autonomous synthesis of (A)** fully-actuated mechanisms and **(B)** under-actuated mechanisms used in assistive exoskeleton robots. Illustrations of the proposed under-actuated mechanism synthesis formulation **(C)** for the DOF evaluation using the reverse work transmittance efficiency, defined as the ratio of output work at the input actuator and input work to the end effector, and **(D)** for displacement-based measurement of the moment direction error.

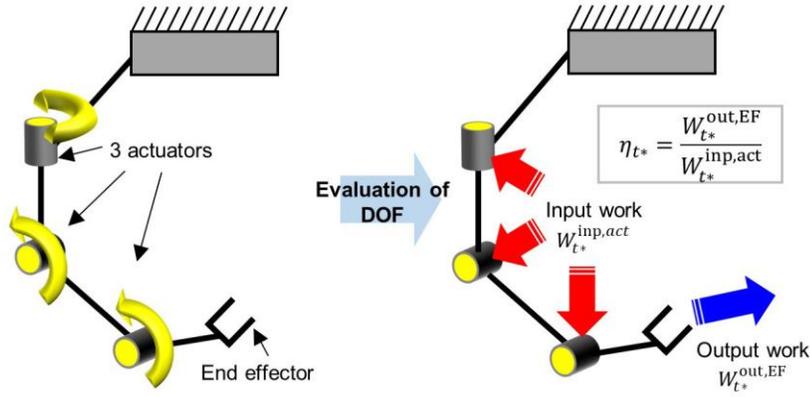

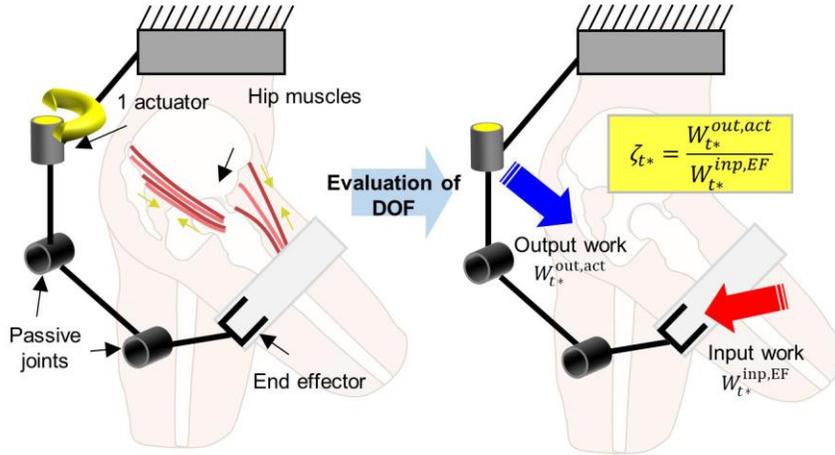

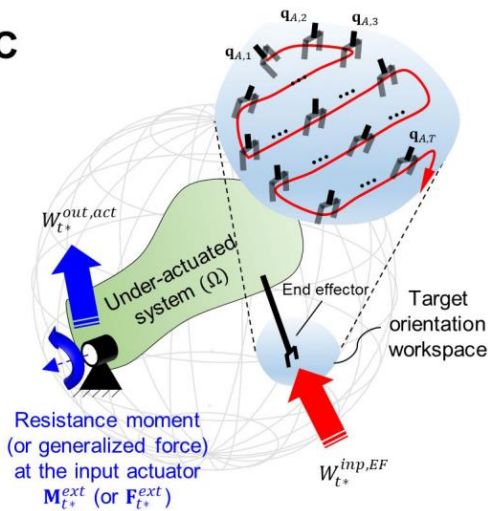
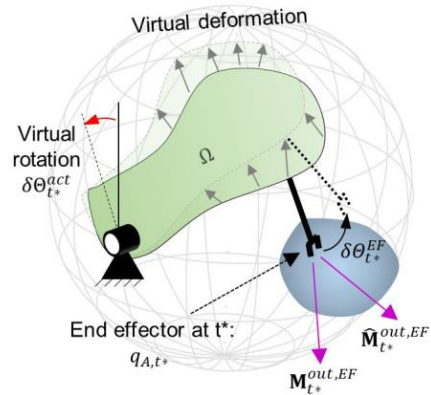

**Fig. 5. Autonomous synthesis results for Case Study 1.** (**A**) Top view of the wearer's movements and configurations of the initial, intermediate, and post-processed spherical-SBM with target output moments (red arrow) and current output moments (blue arrow) at analysis time steps $t^*$=1, 3, 5, and 7. (**B**) Iteration history of the mean value ($\bar{\zeta}$) of the reverse work transmittance efficiency and output moment error values, $\max(\psi_{t^*}^{(1)},\psi_{t^*}^{(2)},\psi_{t^*}^{(3)})$, $\max(\psi_{t^*}^{(4)},\psi_{t^*}^{(5)},\psi_{t^*}^{(6)})$, and $\max(\psi_{t^*}^{(7)},\psi_{t^*}^{(8)},\psi_{t^*}^{(9)})$

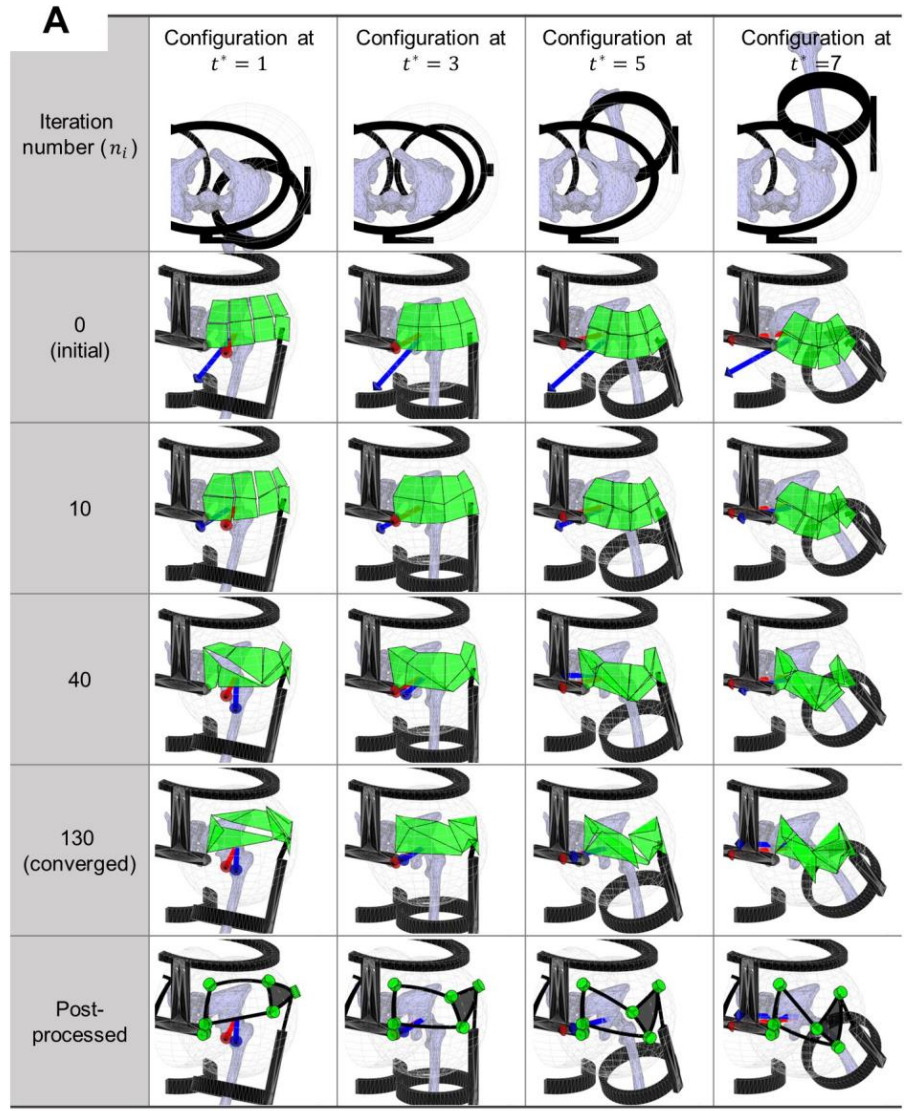

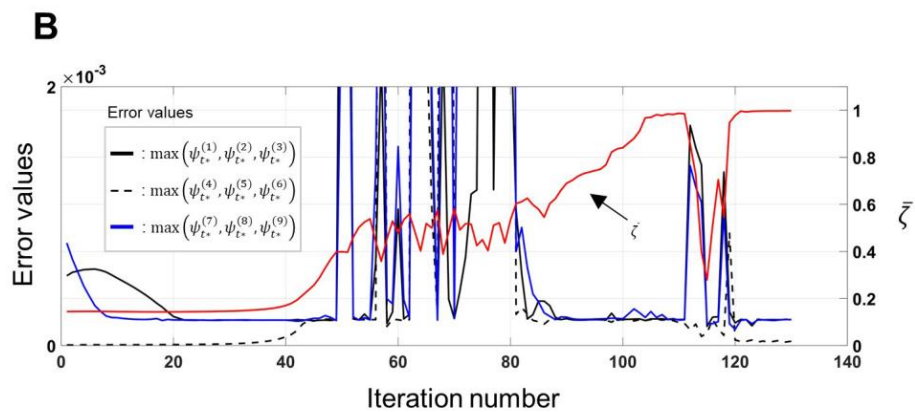

**Fig. 6. The synthesized 'R-4B-R' hip mechanism from ExoSynthesizer. (A)** Illustration of its revolute joints (R1-R6), links (upper harness, L1-L4, thigh harness), screw axes ($\$_1$-$\$_6$) of the joints, and a screw axis ($\$_v$) of its virtual revolute joint. **(B)** The instantaneous rotation axis of the 4-bar mechanism of the synthesized R-4B-R hip mechanism (left) and a serial mechanism, which is instantaneously equivalent to the synthesized mechanism in terms of moment transmission (right). **(C)** Movements of the equivalent serial mechanism defined in (B), with its output moments (or actuation wrenches) (left) and variation of its link dimensions (right) at the target ranges of motion (t*=1~7). **(D)** Top view of the equivalent serial mechanism, its output moments (left), and the desired and synthesized output moment directions (right) at target ranges of motion.

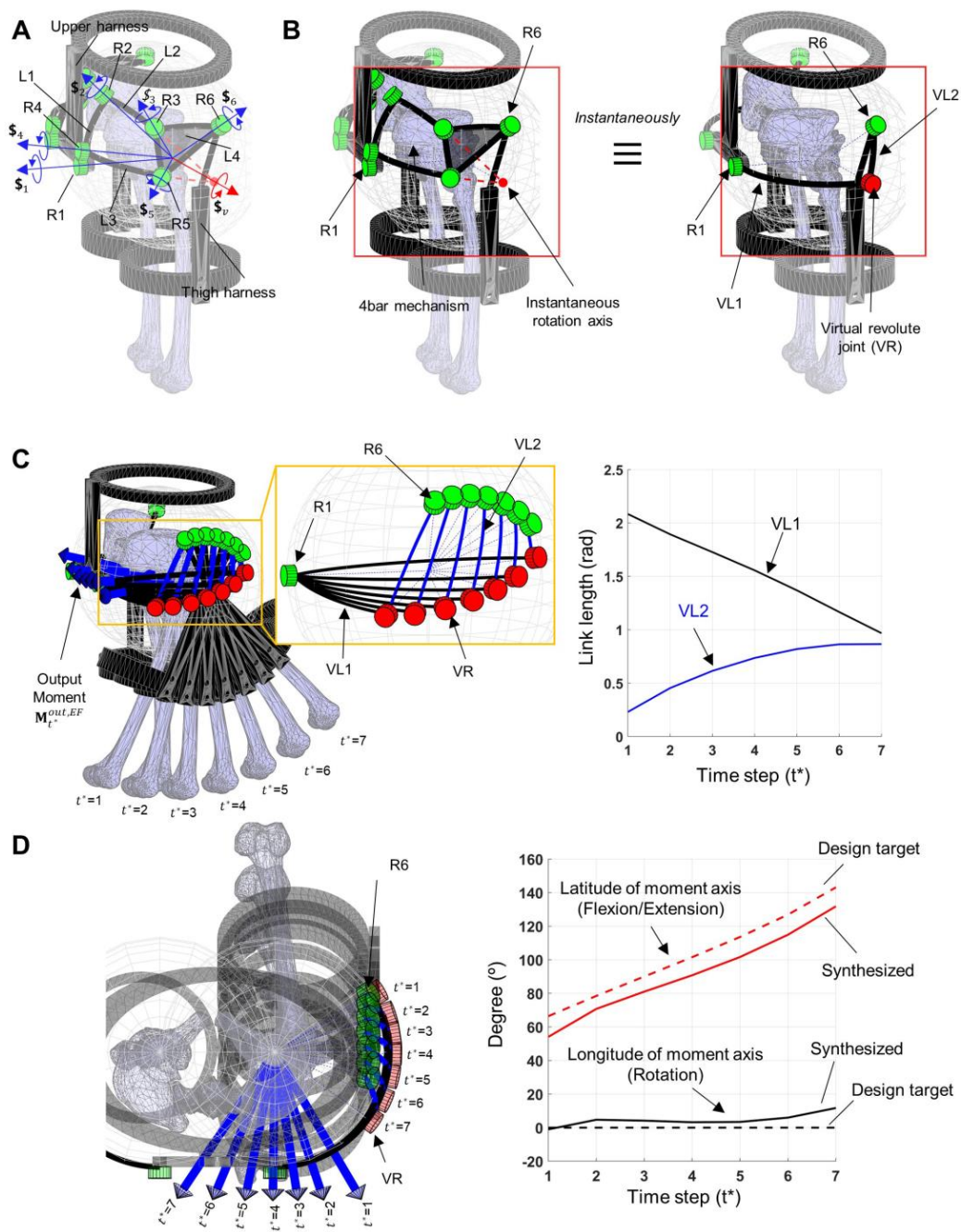

**Fig. 7. Biomechanical assessment of the synthesized R-4B-R mechanism. (A)** Selected three-dimensional output moments that could be generated by right and left R-4B-R mechanisms, resembling the biological hip moments required for normal walking. See Fig. S5 for details. **(B)** Simulated biological hip moments of wearers assisted/unassisted by the output moments in (A) during normal walking by OpenSim. It shows that the hip exoskeleton robot harnessing the R-4B-R mechanism could substantially reduce flexion, extension, and abduction moments while not increasing them in other directions. The moment is supplied only with one actuator per leg, which would be impossible in conventional gait-assistive hip wearable robots requiring more than one actuator per leg.

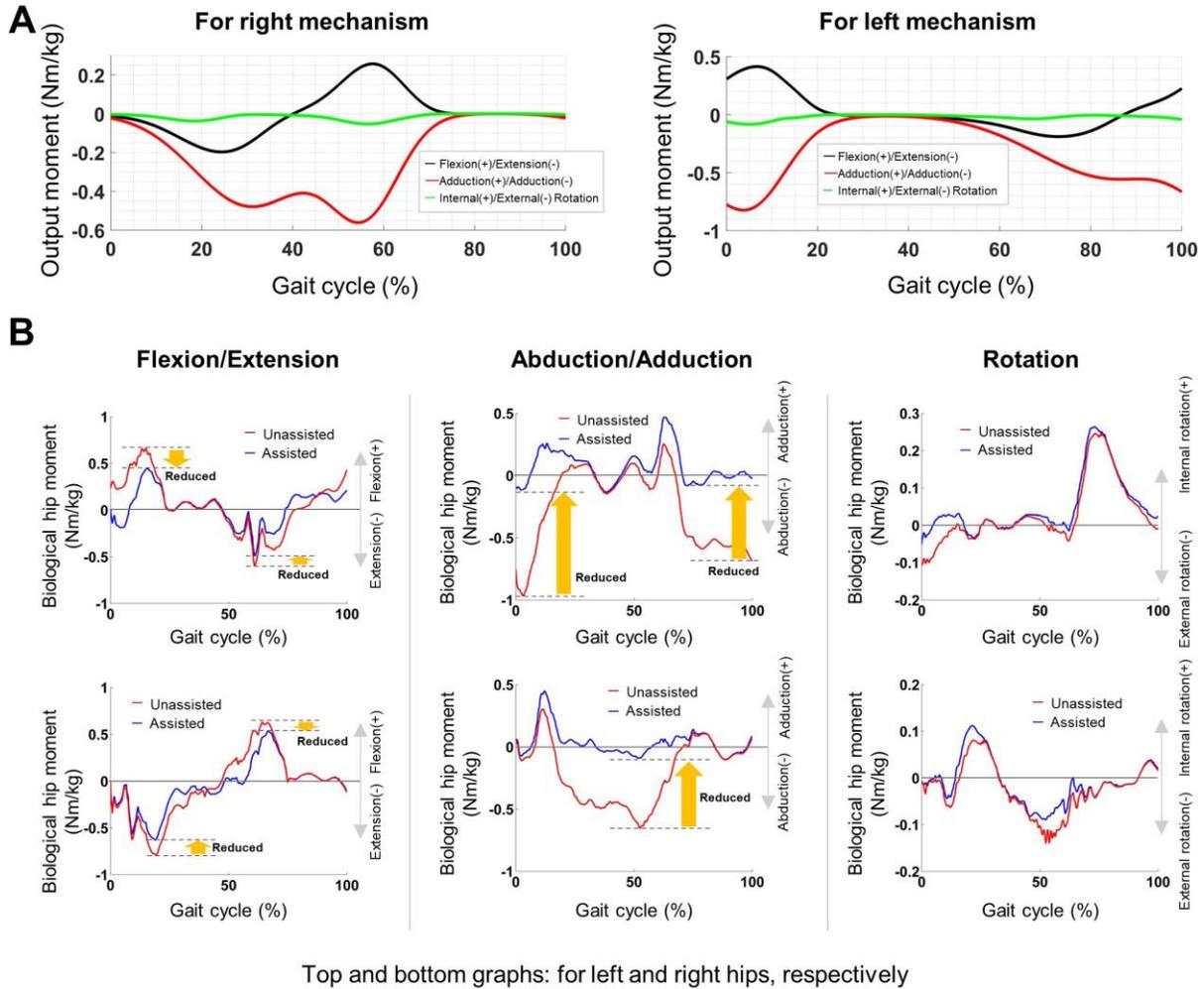

Top and bottom graphs: for left and right hips, respectively

# Supplementary Materials for

# Computational Synthesis of Wearable Robot Mechanisms: Application to Hip-joint Mechanisms

Seok Won *et al.*

Corresponding author: Yoon Young Kim, ykim@snu.ac.kr

**The PDF file includes:**

  Tables S1 to S4

  Figures S1 to S6

  Sections S1 to S6

**Table S1. Target range of motion and output moments for Case Study 2**

| $t^*$ | $(\rho_{A,t^*}, \theta_{A,t^*}, \phi_{A,t^*})$ (rad) | $\widehat{\mathbf{M}}_{t^*}^{out,EF}/M_o$ |
|---|---|---|
| 1 | (0, 0, -0.4) | (0, -1, 0) |
| 2 | (0, 0, -0.2) | (0, -1, 0) |
| 3 | (0, 0, 0) | (0, -1, 0) |
| 4 | (0, 0, 0.2) | (0, -1, 0) |
| 5 | (0, 0, 0.4) | (0, -1, 0) |
| 6 | (0, 0, 0.6) | (0, -1, 0) |
| 7 | (0, 0, 0.8) | (0, -1, 0) |

**Table S2. Target range of motion and output moments for Case Study 3**

| $t^*$ | $(\rho_{A,t^*}, \theta_{A,t^*}, \phi_{A,t^*})$ (rad) | $\widehat{\mathbf{M}}_{t^*}^{out,EF}/M_o$ |
|---|---|---|
| 1 | (0, -0.2, -0.4) | (-0.8, -1, 0) |
| 2 | (0, -0.2, -0.2) | (-0.4, -1, 0) |
| 3 | (0, -0.2, 0) | (0, -1, 0) |
| 4 | (0, -0.2, 0.2) | (0.4, -1, 0) |
| 5 | (0, -0.2, 0.4) | (0.8, -1, 0) |
| 6 | (0, -0.2, 0.6) | (1.2, -1, 0) |
| 7 | (0, -0.2, 0.8) | (1.6, -1, 0) |

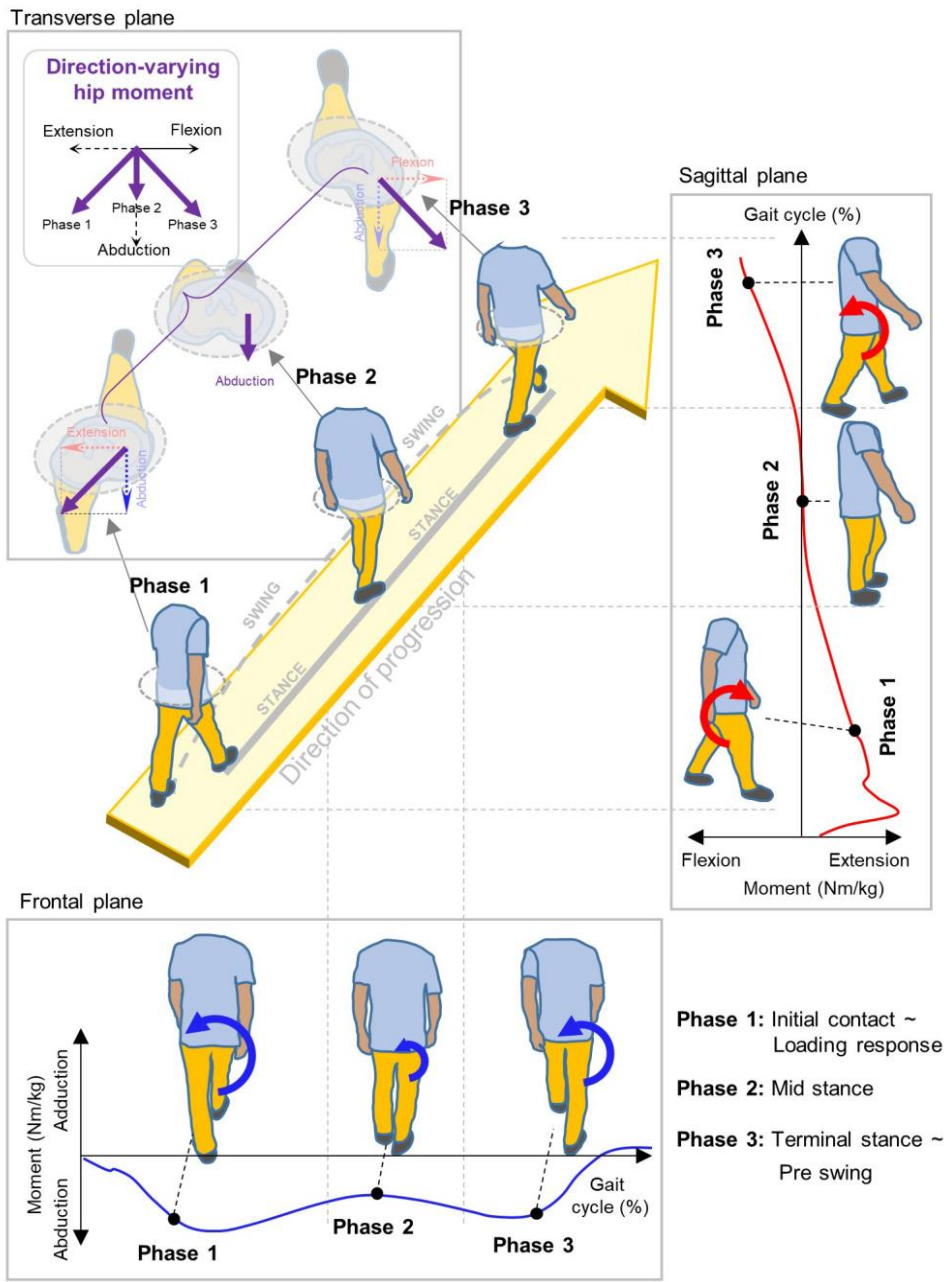

**Fig. S1. Motivation for the development of a computational framework for the design of wearable robot mechanisms.** We can imagine a novel cost-effective gait-assistive hip-exo robot that could offer multidirectional assistance using just one actuator for each hip joint. This would be achievable if a sophisticated mechanism could convert an actuator's input torque into a customized direction-varying moment, aligning with the biological hip moment as depicted in the transverse plane. Currently, neither such a mechanism nor a systematic design approach to synthesize the mechanism has been established.

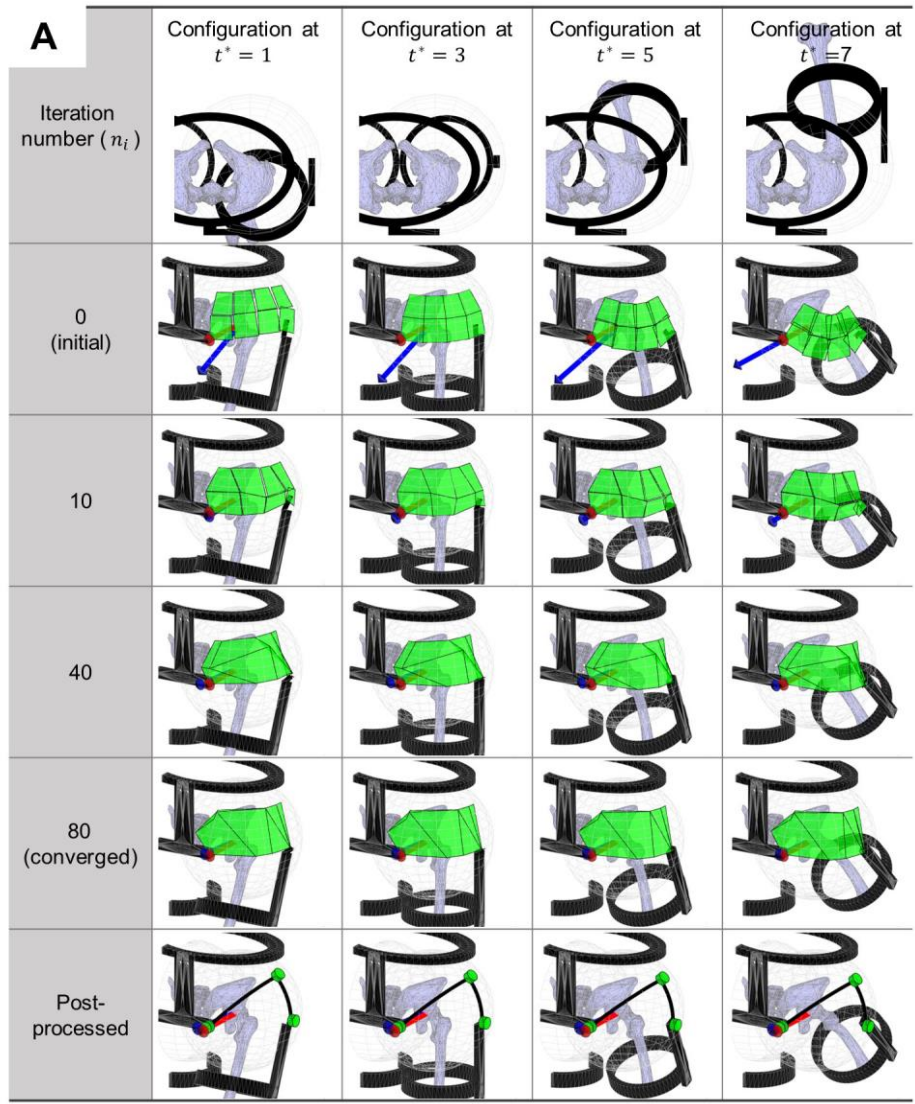

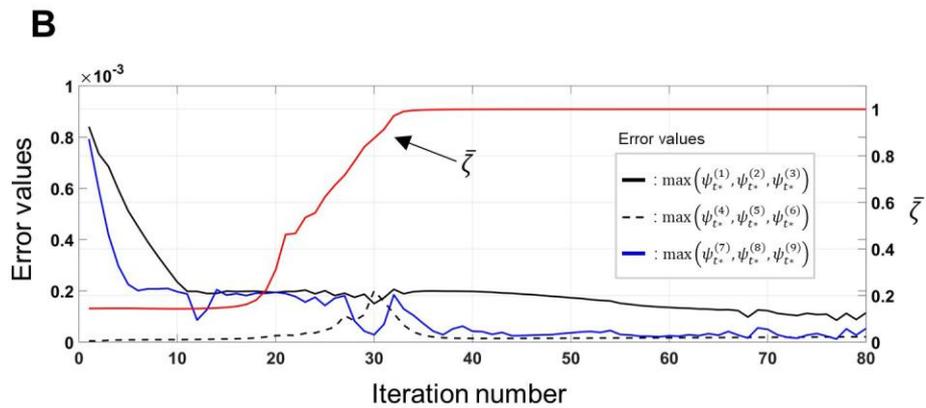

**Fig. S2. Autonomous synthesis for Case Study 2 given in Table S1 resulting in a conventional serial-type mechanism.** (A) Top view of the movement of the wearer and the S-SBM configurations at the initial, intermediate, and converged states for different analysis time steps ($t^*$=1, 3, 5, and 7). Red arrow: target output moment and bule arrow: current output moment. (B) Iteration history of the mean value ($\bar{\zeta}$) of the reverse work transmittance efficiency and output moment error values, $\max(\psi_{t^*}^{(1)}, \psi_{t^*}^{(2)}, \psi_{t^*}^{(3)})$, $\max(\psi_{t^*}^{(4)}, \psi_{t^*}^{(5)}, \psi_{t^*}^{(6)})$, and $\max(\psi_{t^*}^{(7)}, \psi_{t^*}^{(8)}, \psi_{t^*}^{(9)})$.

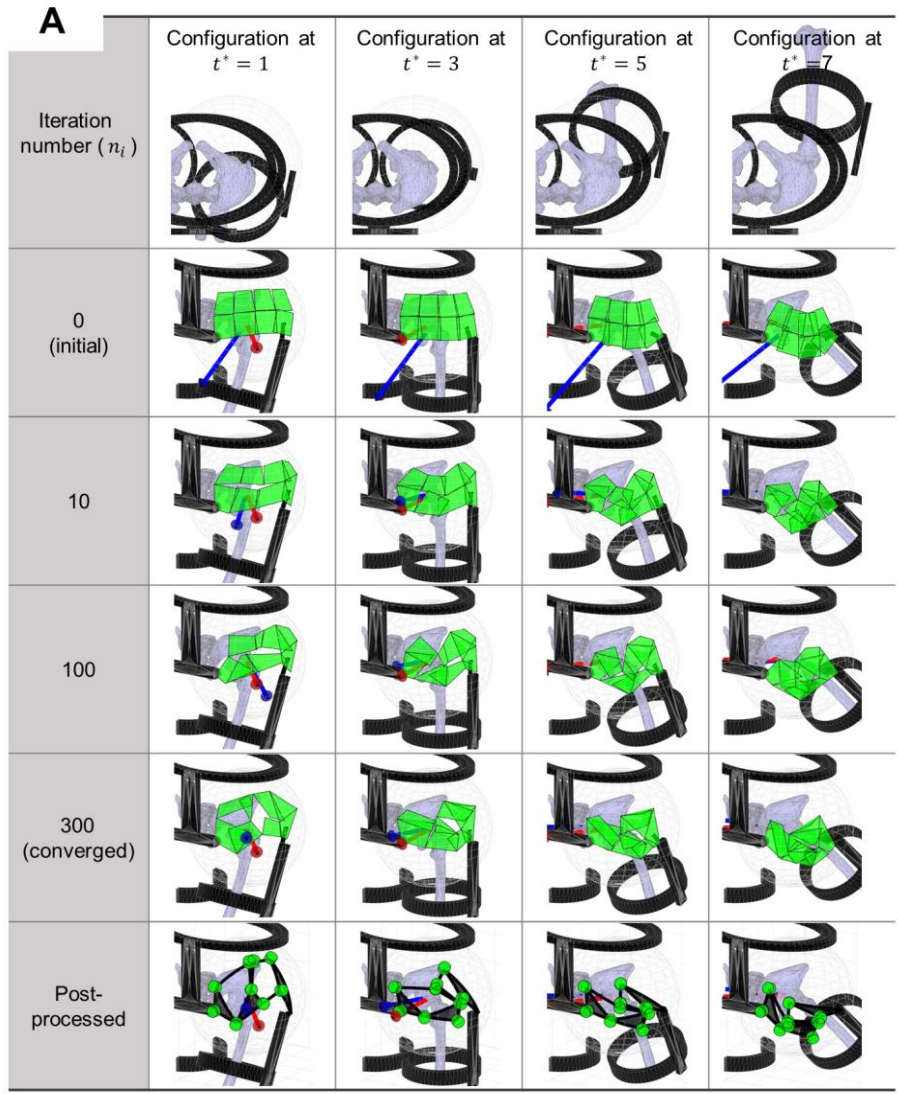

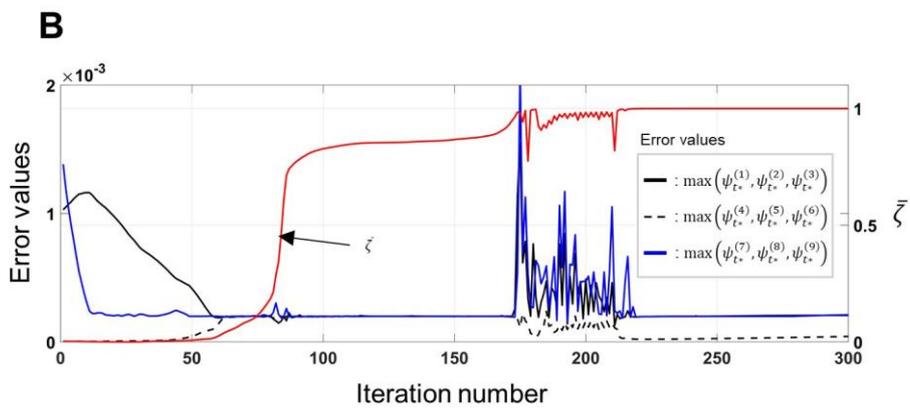

**Fig. S3. Autonomous synthesis for Case Study 3 given in Table S2 resulting in a novel R-4B⊕5B mechanism.** (**A**) Top view of the movement of the wearer and the S-SBM configurations at the initial, intermediate, and converged states for different analysis time steps ($t^*$=1, 3, 5, and 7). Red arrow: target output moment and bule arrow: current output moment. (**B**) Iteration history of mean value ($\bar{\zeta}$) of reverse work transmittance efficiency and output moment error values, $\max(\psi_{t^*}^{(1)}, \psi_{t^*}^{(2)}, \psi_{t^*}^{(3)})$, $\max(\psi_{t^*}^{(4)}, \psi_{t^*}^{(5)}, \psi_{t^*}^{(6)})$, and $\max(\psi_{t^*}^{(7)}, \psi_{t^*}^{(8)}, \psi_{t^*}^{(9)})$.

**Table S3. Geometric specifications of the R-4B-R mechanisms (A)** determined directly from the autonomous synthesis method for Case Study 1 (shown in Fig. 6) and **(B)** modified to perform the biomechanical assessment for a specific wearer (presented in Fig. 7).

| Revolute joint axis | A Before shape modification (obtained from optimization) | B After shape modification |
|:---:|:---:|:---:|
| $r_1$ | (0, -0.9998, 0.0173) | (0,-1,0) |
| $r_2$ | (-0.17, -0.6870, 0.7065) | (0.2248,-0.5034,0.8343) |
| $r_3$ | (0.6971, -0.4213, 0.5802) | (0.7030,-0.3335,0.6282) |
| $r_4$ | (0.0008, -0.9845, 0.1754) | (0.7447,-0.6130,0.2639) |
| $r_5$ | (0.9040, -0.4129, 0.1114) | (0.9589,-0.1888,0.2117) |
| $r_6$ | (0.7749, 0.2819, 0.5657) | (0.5794,-0.0309,0.8145) |

## S1. Kinematic analysis using the spherical-SBM

The motion of a mechanism represented by the spherical-SBM (S-SBM) cannot be directly analyzed using conventional kinematics because the S-SBM involves elastic deformations of zero-length springs connecting rigid spherical blocks. Therefore, its kinematic behavior is alternatively analyzed using the static force equilibrium of the S-SBM. To derive the equilibrium equation, the total strain energy stored in all zero-length springs is used.

Let us consider the $m^{\text{th}}$ zero-length spring (having spring stiffness $k_m$) as shown in Fig. S4. The spring is assumed to deform linearly regardless of the magnitude of its displacement. The adjacent two blocks connected to the spring are denoted by Block (*m*, 1) and Block (*m*, 2) ($1 \leq m \leq N_S$), where $N_S$ is the total number of zero-length springs used in forming the spherical-SBM (Fig. S4A). These blocks will also be represented by Block *n* ($1 \leq n \leq N_B$) as long as notation *n* and notation (*m,j*) are properly related. In the spherical-SBM used in this study, all blocks are assumed to be connected by ball joints about the hip joint center *O* (although the joints are not illustrated in Fig. S4). The deformed state of each block can be represented by three rotational state variables $(\rho, \theta, \phi)$, which are the Tait-Bryan angles discussed in the main text.

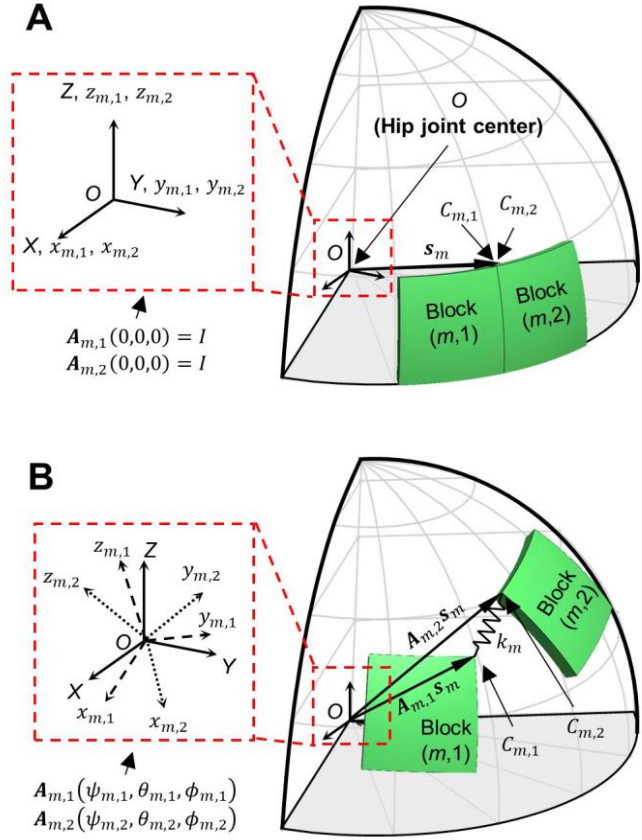

**Fig. S4.** **(A)** Undeformed and **(B)** deformed configurations of the $m^{th}$ zero-length spring and two blocks connected to it

The strain energy $U_m$ stored in the $m^{th}$ spring can be expressed as

$$U_m = \frac{1}{2} k_m \left( \mathbf{u}_m \right)^T \mathbf{u}_m, \tag{S1}$$

where $\mathbf{u}_m$ denotes the relative displacement of the corner node $C_{m,2}$ with respect to the corner node $C_{m,1}$ (see Fig. S4B). The displacement $\mathbf{u}_m$ can be calculated using the following expression:

$$\mathbf{u}_m = \mathbf{A}_{m,2} \mathbf{s}_m - \mathbf{A}_{m,1} \mathbf{s}_m, \tag{S2}$$

where

$$\mathbf{A}_{m,i}\left(\rho_{m,i},\theta_{m,i},\phi_{m,i}\right)$$

$$=\begin{bmatrix} \cos\rho & 0 & \sin\rho \\ 0 & 1 & 0 \\ -\sin\rho & 0 & \cos\rho \end{bmatrix}_{m,i} \begin{bmatrix} \cos\theta & -\sin\theta & 0 \\ \sin\theta & \cos\theta & 0 \\ 0 & 0 & 1 \end{bmatrix}_{m,i} \begin{bmatrix} 1 & 0 & 0 \\ 0 & \cos\phi & -\sin\phi \\ 0 & \sin\phi & \cos\phi \end{bmatrix}_{m,i} \quad (i=1,2), \quad (S3)$$

$$=\begin{bmatrix} \cos\rho\cos\theta & \sin\rho\sin\phi - \cos\rho\cos\phi\sin\theta & \cos\phi\sin\rho + \cos\rho\sin\theta\sin\phi \\ \sin\theta & \cos\theta\cos\phi & -\cos\theta\sin\phi \\ -\cos\theta\sin\rho & \cos\rho\sin\phi + \cos\phi\sin\rho\sin\theta & \cos\rho\cos\phi - \sin\rho\sin\theta\sin\phi \end{bmatrix}_{m,i}$$

and

$$\mathbf{s}_m\left(\Theta_l,\Phi_l\right)=\begin{bmatrix} \sin\Theta_l\cos\Phi_l \\ \sin\Theta_l\sin\Phi_l \\ \cos\Theta_l \end{bmatrix}. \quad (S4)$$

In Eqs. S3 and S4, $\mathbf{A}_{m,i}$ is the three-dimensional rotation matrix defining the current state of the body-fixed coordinate system ($x_{m,i}$, $y_{m,i}$, $z_{m,i}$) of Block ($m,i$) ($i=1,2$) relative to the global coordinate system ($X$, $Y$, $Z$) centered at $O$. Initially, the body-fixed coordinate system is set to coincide with the global coordinate system. The symbol $\mathbf{s}_m$ denotes the position vector of the corner nodes $C_{m,1}$ and $C_{m,2}$ at the initial undeformed state.

Besides the zero-length block-connecting springs, there is an anchoring spring used to connect Block $I$ (=1) to the ground. The input actuator is regarded to be connected to Block $I$ (=1) in Fig. 3A. If the node representing the input actuator location is denoted by $C_o$, the stored energy in the corresponding anchoring spring is given by

$$U_o = \frac{1}{2}k_{max}\left(\mathbf{u}_o\right)^T \mathbf{u}_o, \quad (S5)$$

where $k_{max}$ is used for spring stiffness because this spring with $k = k_{max}$ simulates a revolute joint connecting the ground and Block 1. The displacement $\mathbf{u}_o$ is expressed as

$$\mathbf{u}_o = \mathbf{A}_o \mathbf{s}_o - \mathbf{s}_o, \quad (S6)$$

where matrix $\mathbf{A}_o$ denotes the rotation matrix for Block 1, and vector $\mathbf{s}_o$ represents the position vector of corner node $C_0$ at the initial state, respectively. If we introduce the symbol $t^*$ to denote a specific time step during the motion of the mechanism being synthesized, the total strain energy stored in all springs at time step $t^*$ can be expressed as

$$U_{t^*} = \sum_{m=1}^{N_S} U_{m,t^*} + U_{o,t^*}, \quad (S7)$$

where $U_{m,t^*}$ and $U_{o,t^*}$ denote the strain energy stored in the $m^{th}$ spring and the anchoring spring at time step $t^*$, respectively.

Once the expression for the system strain energy $U_{t^*}$ is found, the force equilibrium equation for the system of spherical rigid blocks is derived as

$$\mathbf{F}_{t^*}^{int}(\mathbf{v}_{t^*}) = \mathbf{F}_{t^*}^{ext} \quad \text{for given} \quad \mathbf{q}_{A,t^*} \ (t^* = 1, 2, \cdots, T). \tag{S8}$$

In Eq. S8,

$\mathbf{F}_{t^*}^{int} = \dfrac{dU_{t^*}}{d\mathbf{v}_{t^*}}$ : the internal force in the system at time step $t^*$,

$\mathbf{F}_{t^*}^{ext}$ : the external force applied to the system (as used in defining the reverse work transmission efficiency) at time step $t^*$,

$\mathbf{q}_{A,t^*} = [\rho_{A,t^*}, \theta_{A,t^*}, \phi_{A,t^*}]^T$ : input state variable vector given to block $A$ ($A=4$ in the case of Figure 2A) at time step $t^*$, consisting of the Tait-Bryan angles of the block,

$\mathbf{v}_{t^*} = \left[ (\mathbf{q}_{1,t^*})^T, (\mathbf{q}_{2,t^*})^T, \cdots, (\mathbf{q}_{(A-1),t^*})^T, (\mathbf{q}_{(A+1),t^*})^T, \cdots, (\mathbf{q}_{N_B,t^*})^T \right]^T$ : a set of state variable vectors $\mathbf{q}_{n,t^*} = [\rho_{n,t^*}, \theta_{n,t^*}, \phi_{n,t^*}]^T$ ( $n = 1, 2, \cdots (A-1), (A+1), \cdots N_B$ ), where $\rho_{n,t^*}$, $\theta_{n,t^*}$, and $\phi_{n,t^*}$ are the Tait-Bryan angles of Block $n$ at time step $t^*$, (See Eq. S3 for the definition of the angles)

$T$: total number of time steps used for the analysis.

The internal force vector in Eq. S8 can be expressed as

$$\dfrac{dU_{t^*}}{d\mathbf{v}_{t^*}} = \left\{ \left( \dfrac{dU}{d\mathbf{q}_1} \right)^T, \left( \dfrac{dU}{d\mathbf{q}_2} \right)^T, \cdots, \left( \dfrac{dU}{d\mathbf{q}_{(A-1)}} \right)^T, \left( \dfrac{dU}{d\mathbf{q}_{(A+1)}} \right)^T, \cdots, \left( \dfrac{dU}{d\mathbf{q}_{N_B}} \right)^T \right\}^T \bigg|_{t^*}, \tag{S9}$$

where

$$\dfrac{dU_{t^*}}{d\mathbf{q}_{n,t^*}} = \left\{ \dfrac{dU_{t^*}}{d\rho_{n,t^*}}, \dfrac{dU_{t^*}}{d\theta_{n,t^*}}, \dfrac{dU_{t^*}}{d\phi_{n,t^*}} \right\}^T \quad (\text{for} \ n = 1, 2, \cdots, (A-1), (A+1), \cdots, N_B), \tag{S10}$$

with

$$\dfrac{dU_{t^*}}{d\rho_{n,t^*}} = \sum_{i \in S_n} k_i \left( \dfrac{d\mathbf{A}_{i,2,t^*}}{d\rho_{n,t^*}} \mathbf{s}_{i,t^*} \right)^T \mathbf{u}_{i,t^*}, \tag{S11}$$

$$\dfrac{dU_{t^*}}{d\theta_{n,t^*}} = \sum_{i \in S_n} k_i \left( \dfrac{d\mathbf{A}_{i,2,t^*}}{d\theta_{n,t^*}} \mathbf{s}_{i,t^*} \right)^T \mathbf{u}_{i,t^*}, \tag{S12}$$

$$\dfrac{dU_{t^*}}{d\phi_{n,t^*}} = \sum_{i \in S_n} k_i \left( \dfrac{d\mathbf{A}_{i,2,t^*}}{d\phi_{n,t^*}} \mathbf{s}_{i,t^*} \right)^T \mathbf{u}_{i,t^*}. \tag{S13}$$

The set $S_n$ in Eqs. S11-S13 is defined as

$S_n = \{ m \mid \text{indices denoting zero-length springs attached to block } n \}$.

To solve Eq. S8, a numerical solver, 'fsolve,' in MATLAB was used. It requires $d\mathbf{F}_{t^*}^{ext}/d\mathbf{v}_{t^*}$ and $d\mathbf{F}_{t^*}^{int}/d\mathbf{v}_{t^*}$, which is the Jacobian matrix $\mathbf{J}_{t^*}$ of the system. Since these expressions are easy to derive and lengthy, they are not explitly given here.

## S2. Setup for the evaluation of the reverse work transmittance efficiency

To calculate the mean value of the reverse work transmittance efficiency appearing in Eq. 6, we need to define the resistive applied moments and the input displacement to the mechanism being synthesized elaborately. In gait-assistive wearable hip-exo robots, 3-DOF rotational displacements should be considered input displacements. To this end, we first determine the target orientation workspace (or range of motion) of the end-effector to cover all hip angles corresponding to the representative gait motions as prescribed in Fig. 2B for Case Study 1 and Tables S1 and S2 for Case Studies 2 and 3, respectively. The workspace is sketched on the left side of Fig. 4C. We then discretize the orientation workspaces using equally-distributed points (or spatial poses in this study) and define a fictitious trajectory that sequentially passes through the points (or the poses) to which the analysis time steps from $t^*=1$ to $t^*=N$ are assigned according to the passing order.

To be able to calculate the efficiency function, we introduce the external moment[1] applied to the input actuator in the following form:

$$\mathbf{M}_{t^*}^{ext} = F_0 \left( \mathbf{w}_{t^*} \times \mathbf{w}_{t^*-1} \right) \; \left( t^*=1,2,\cdots,T \right), \tag{S14}$$

where $\mathbf{w}_{t^*}$ is the position vector of a specified point in Block $I$ ($=1$) to which the input actuator is attached at time step $t^*$ and $F_0$ is a constant value. The vector $\mathbf{w}_{t^*}$ is defined as

$$\mathbf{w}_{t^*} = \mathbf{A}_{I,t^*} [0,0,1]^T, \tag{S15}$$

where $\mathbf{A}_{I,t^*}$ represents the rotation matrix of Block $I$ defined in Eq. S3. Because $\mathbf{M}_{t^*}^{ext}$ is orthogonal to $\mathbf{w}_{t^*}$ and $\mathbf{w}_{t^*-1}$ (see Eq. S14), $\mathbf{M}_{t^*}^{ext}$ is applied precisely against the rotational motion of Block $I$ at every time step. As the end-effector is set to follow the fictitious trajectory within the target orientation workspace, the moment $\mathbf{M}_{t^*}^{ext}$ produces positive external work $W_{t^*}^{out,act}$ at the actuator, as given by Eq. 10. If the strain energy $U_{t^*}$ is calculated using Eq. S7, the reverse work transmittance efficiency $\zeta_{t^*}$ in Eq. 9 can be calculated. Then the time-averaged reverse work transmittance efficiency $\overline{\zeta}$ appearing in Eq. 6 can be evaluated.

Referring to Eq. 10, the external work is expressed by $\mathbf{F}_t^{ext}$, not by $\mathbf{M}_t^{ext}$. Because the states of all rigid spherical blocks, including Block $I$, are expressed in terms of the Tait-Bryan angle, the applied moment should be expressed in a generalized force form for this angle. Thus, $\mathbf{M}_t^{ext}$ is converted to $\mathbf{F}_t^{ext}$ by the following transformation:

$$\mathbf{F}_t^{ext} = \left[ \mathbf{R}_{I,t}^y \mathbf{e}_y, \; \mathbf{R}_{I,t}^y \mathbf{R}_{I,t}^z \mathbf{e}_z, \; \mathbf{R}_{I,t}^y \mathbf{R}_{I,t}^z \mathbf{R}_{I,t}^x \mathbf{e}_x \right]^T \mathbf{M}_t^{ext}, \tag{S16}$$

where the rotation matrices $\mathbf{R}_{I,t}^y$, $\mathbf{R}_{I,t}^z$, and $\mathbf{R}_{I,t}^x$ are given by

$$\mathbf{R}_{I,t}^y = \begin{bmatrix} \cos \rho_{I,t} & 0 & \sin \rho_{I,t} \\ 0 & 1 & 0 \\ -\sin \rho_{I,t} & 0 & \cos \rho_{I,t} \end{bmatrix}, \tag{S17}$$

---

[1] Note that this moment is considered only to calculate the reverse work transmission efficiency and is not the actual moment applied to the input actuator.

$$\mathbf{R}_{I,t}^z = \begin{bmatrix} \cos\theta_{I,t} & -\sin\theta_{I,t} & 0 \\ \sin\theta_{I,t} & \cos\theta_{I,t} & 0 \\ 0 & 0 & 1 \end{bmatrix}, \tag{S18}$$

$$\mathbf{R}_{I,t}^x = \begin{bmatrix} 1 & 0 & 0 \\ 0 & \cos\phi_{I,t} & -\sin\phi_{I,t} \\ 0 & \sin\phi_{I,t} & \cos\phi_{I,t} \end{bmatrix}, \tag{S19}$$

and the unit vectors $\mathbf{e}_y$, $\mathbf{e}_z$, and $\mathbf{e}_x$ are defined as

$$\mathbf{e}_y = \begin{bmatrix} 0 \\ 1 \\ 0 \end{bmatrix}, \mathbf{e}_z = \begin{bmatrix} 0 \\ 0 \\ 1 \end{bmatrix}, \mathbf{e}_x = \begin{bmatrix} 1 \\ 0 \\ 0 \end{bmatrix}. \tag{S20}$$

## S3. Constraint equations

We will explain how to calculate the measurements $\left[\Delta(\mathbf{p}_{j,t^*})_X, \Delta(\mathbf{p}_{j,t^*})_Y, \Delta(\mathbf{p}_{j,t^*})_Z\right]_{j=1,2,3}$ and set their target values $\left[\Delta(\mathbf{p}_{j,t^*})_X, \Delta(\mathbf{p}_{j,t^*})_Y, \Delta(\mathbf{p}_{j,t^*})_Z\right]_{j=1,2,3}$ in constraint equation 14.

The vectors $\Delta(\mathbf{p}_{1,t^*})_\alpha$, $\Delta(\mathbf{p}_{2,t^*})_\alpha$, and $\Delta(\mathbf{p}_{3,t^*})_\alpha$ (for $\alpha = X, Y, Z$) denote variations of the points $\mathbf{p}_{1,t^*} = [1,0,0]^T$, $\mathbf{p}_{2,t^*} = [0,1,0]^T$, and $\mathbf{p}_{3,t^*} = [0,0,1]^T$ of Block $I$ (the block to which the input actuator is connected), respectively, when the end effector (or Block $A$) is rotated around the global $\alpha$-axis through a small perturbation angle $\Delta\theta_\alpha$ (=$10^{-3}$ in this study) with respect to the state $\mathbf{q}_{A,t^*}$ (input state at time $t^*$). Figure 4D illustrates this situation. The transformation matrix for the end effector after the perturbation, $\tilde{\mathbf{A}}_{\alpha,t^*}(\tilde{\mathbf{q}}_{A,\alpha,t^*})$, can be calculated by

$$\tilde{\mathbf{A}}_{\alpha,t^*}(\tilde{\mathbf{q}}_{A,\alpha,t^*}) = \Delta\mathbf{A}_\alpha \mathbf{A}_{A,t^*}(\mathbf{q}_{A,t^*}) \tag{S21}$$

with

$$\Delta\mathbf{A}_X = \begin{bmatrix} 1 & 0 & 0 \\ 0 & \cos\Delta\theta_X & -\sin\Delta\theta_X \\ 0 & \sin\Delta\theta_X & \cos\Delta\theta_X \end{bmatrix}, \tag{S22}$$

$$\Delta\mathbf{A}_Y = \begin{bmatrix} \cos\Delta\theta_Y & 0 & \sin\Delta\theta_Y \\ 0 & 1 & 0 \\ -\sin\Delta\theta_Y & 0 & \cos\Delta\theta_Y \end{bmatrix}, \tag{S23}$$

$$\Delta\mathbf{A}_Z = \begin{bmatrix} \cos\Delta\theta_Z & -\sin\Delta\theta_Z & 0 \\ \sin\Delta\theta_Z & \cos\Delta\theta_Z & 0 \\ 0 & 0 & 1 \end{bmatrix}, \tag{S24}$$

where $\Delta\mathbf{A}_\alpha$ and $\mathbf{A}_{A,t^*}$ are the transformation matrix for the small rotation around the global $\alpha$-axis and the orientation matrix of Block $A$ before the rotation, respectively. By substituting $\tilde{\mathbf{q}}_{A,\alpha,t^*}$ obtained from Eq. S21 into Eq. S8, we can obtain $\tilde{\mathbf{v}}_{\alpha,t^*}$, the perturbed state variables of $\mathbf{v}_{t^*}$. In solving Eq. S8, it should be noted that the resistance moment $\tilde{\mathbf{M}}_{\alpha,t^*}^{ext} = F_0(\tilde{\mathbf{w}}_{\alpha,t^*} \times \mathbf{w}_{t^*-1})$ is exerted on Block $I$ as an external force $\mathbf{F}_{t^*}^{ext}$, where $\tilde{\mathbf{w}}_{\alpha,t^*} = \tilde{\mathbf{A}}_{\alpha,t^*}[0,0,1]^T$ and $\mathbf{w}_{t^*-1} = \mathbf{A}_{I,t^*-1}[0,0,1]^T$ as given in Eqs. S14 and S15. By Chasles' theorem (45), $\tilde{\mathbf{q}}_{I,\alpha,t^*}$, the perturbed state of Block $I$, which can be obtained from $\tilde{\mathbf{v}}_{\alpha,t^*}$, can be expressed as a rotation about an axis from $\mathbf{q}_{I,t^*}$, the non-perturbed state of the block, which can be obtained from $\mathbf{v}_{t^*}$. Let us define the rotation axis as $\boldsymbol{\omega}_{\alpha,t^*}$ ($\|\boldsymbol{\omega}_{\alpha,t^*}\|=1$) and the rotation angle as $\Delta\lambda_{\alpha,t^*}$. Then, the position variation vectors $\Delta(\mathbf{p}_{1,t^*})_\alpha$, $\Delta(\mathbf{p}_{2,t^*})_\alpha$, and $\Delta(\mathbf{p}_{3,t^*})_\alpha$ can be written as:

$$\Delta(\mathbf{p}_{j,t^*})_\alpha = \{\exp(\Delta\lambda_{\alpha,t^*}\mathbf{W}_{\alpha,t^*}) - \mathbf{I}\}\mathbf{p}_{j,t^*} \quad \text{(for } j=1,2,3\text{)}, \tag{S25}$$

where $\mathbf{W}_{\alpha,t^*}$ is the cross product matrix of $\boldsymbol{\omega}_{\alpha,t^*}$ ($\mathbf{W}_{\alpha,t^*} = [\boldsymbol{\omega}_{\alpha,t^*}]_\times$) and $\mathbf{I}$ is the $3\times 3$ identity matrix.

Now, let us explain how the target position variation vectors $\left[\Delta(\mathbf{p}_{j,t^*})_X, \Delta(\mathbf{p}_{j,t^*})_Y, \Delta(\mathbf{p}_{j,t^*})_Z\right]_{j=1,2,3}$ could be determined. In the current hip-exo synthesis problems, the input actuator moment is $M_0[0 \ -1 \ 0]^T$ and the target output moments are given as $\hat{\mathbf{M}}_{t^*}^{out,EF} = M_0[\kappa_X, \kappa_Y = -1, \kappa_Z = 0]^T$. From the principle of virtual work stated in Eq. 13, the target perturbation amount for Block $I$ with respect to its fixed rotation axis $\boldsymbol{\omega}_{\alpha,t^*} = [0,-1,0]^T$ is $\Delta\lambda_{\alpha,t^*} = \kappa_\alpha \Delta\theta_\alpha = \kappa_\alpha \times 10^{-3}$ ($\alpha = X,Y,Z$) when Block $A$ (or end effector) is rotated around the global $\alpha$-axis through a small perturbation angle $\Delta\theta_\alpha = 10^{-3}$ Therefore, the target vectors $\left[\Delta(\mathbf{p}_{j,t^*})_X, \Delta(\mathbf{p}_{j,t^*})_Y, \Delta(\mathbf{p}_{j,t^*})_Z\right]_{j=1,2,3}$ can be given by

$$\Delta\left(\mathbf{p}_{j,t^*}\right)_\alpha = \left\{\exp\left(\kappa_\alpha \times 10^{-3} \times \begin{bmatrix} 0 & 0 & -1 \\ 0 & 0 & 0 \\ 1 & 0 & 0 \end{bmatrix}\right) - \mathbf{I}\right\}\mathbf{p}_{j,t^*} \quad \text{(for } j=1,2,3 \text{ and } \alpha = X,Y,Z\text{). (S26)}$$

## S4. Design sensitivity analysis for the gradient-based design optimization

To use a gradient-based optimization algorithm to solve the problem set up by Eqs. 6 and 7, the sensitivities of the objective and constraint functions with respect to the design variable $\xi$ are needed.

The sensitivity of the efficiency function $\bar{\zeta}$ with respect to $\xi$ is

$$\frac{d\bar{\zeta}}{d\xi} = \frac{\partial \bar{\zeta}}{\partial \xi} + \sum_{t^*=1}^{T}\left(\frac{d\mathbf{v}_{t^*}}{d\xi}\frac{\partial \bar{\zeta}}{\partial \mathbf{v}_{t^*}}\right). \tag{S27}$$

The matrix $d\mathbf{v}_{t^*}/d\xi$ in Eq. S27 can be found by directly differentiating the equilibrium equation S8:

$$\frac{d\left(\mathbf{F}_{t^*}^{\text{int}} - \mathbf{F}_{t^*}^{\text{ext}}\right)}{d\xi} = \frac{\partial \mathbf{F}_{t^*}^{\text{int}}}{\partial \xi} + \frac{d\mathbf{v}_{t^*}}{d\xi}\frac{\partial \left(\mathbf{F}_{t^*}^{\text{int}} - \mathbf{F}_{t^*}^{\text{ext}}\right)}{\partial \mathbf{v}_{t^*}} = \frac{\partial \mathbf{F}_{t^*}^{\text{int}}}{\partial \xi} + \frac{d\mathbf{v}_{t^*}}{d\xi}\left(\mathbf{J}_{t^*} - \frac{\partial \mathbf{F}_{t^*}^{\text{ext}}}{\partial \mathbf{v}_{t^*}}\right) = \mathbf{0} \tag{S28}$$

and

$$\frac{d\mathbf{v}_{t^*}}{d\xi} = -\frac{\partial \mathbf{F}_{t^*}^{\text{int}}}{\partial \xi}\left(\mathbf{J}_{t^*} - \frac{\partial \mathbf{F}_{t^*}^{\text{ext}}}{\partial \mathbf{v}_{t^*}}\right)^{-1}, \tag{S29}$$

where $\mathbf{J}_{t^*}$ is the Jacobian matrix of the S-SBM system at analysis time step $t^*$. Since all vectors and matrices appearing in Eqs. S28 and S29 are either already calculated in the previous sections or can be calculated through simple calculations, we will skip giving the detailed expression for $d\bar{\zeta}/d\xi$.

Meanwhile, the sensitivity of the constraint functions, $d\psi_{t^*}^{(i)}/d\xi$, can be obtained using the sensitivities of the vectors $\Delta\left(\mathbf{p}_{j,t^*}\right)_\alpha$ (for j=1,2,3 and $\alpha = X, Y, Z$). Because they denote position variations of the fixed points, $\mathbf{p}_{1,t^*} = [1,0,0]^T$, $\mathbf{p}_{2,t^*} = [0,1,0]^T$, and $\mathbf{p}_{3,t^*} = [0,0,1]^T$, which are independent of the design variables, only the design sensitivities of the points for perturbed states, $\tilde{\mathbf{v}}_{X,t^*}$, $\tilde{\mathbf{v}}_{Y,t^*}$, and $\tilde{\mathbf{v}}_{Z,t^*}$, which are explained in Section S3, need to be considered. They can be obtained from the direct differentiation of the force equilibrium, Eq. S8, as follows:

$$\frac{d\tilde{\mathbf{v}}_{\alpha,t^*}}{d\xi} = -\frac{\partial \tilde{\mathbf{F}}_{\alpha,t^*}^{\text{int}}}{\partial \xi}\left(\tilde{\mathbf{J}}_{\alpha,t^*} - \frac{\partial \tilde{\mathbf{F}}_{\alpha,t^*}^{\text{ext}}}{\partial \mathbf{v}_{\alpha,t^*}}\right)^{-1} \text{ (for } \alpha = X, Y, Z), \tag{S30}$$

where the vectors $\tilde{\mathbf{F}}_{\alpha,t^*}^{\text{int}}$, $\tilde{\mathbf{F}}_{\alpha,t^*}^{\text{ext}}$, and matrix $\tilde{\mathbf{J}}_{\alpha,t^*}^{\text{ext}}$ are the internal force vector, the external force vector, and the Jacobian matrix of the S-SBM system perturbed by a slight rotation of the end effector with respect to the global $\alpha$-axis from its initial state at the time step $t^*$, respectively.

# S5. Description of a selected gait motion and the input actuator torque profile for the R-4B-R mechanism-equipped hip-exo robot

Figure S5 shows the sample gait motion, the output moment direction profiles generated by a modified R-4B-R mechanism according to the sample gait motion, and the selected input torque profiles. First, Figure S5A shows the hip angles for the sample gait motion across the gait cycle:

- flexion/extension angles for the right/left hip joint: denoted by Flex. (+)/Ext. (-) (right/left),

- adduction/abduction angles for the right/left hip joint: denoted by Add. (+)/Abd. (-) (right/left),

- internal/external rotation angles for the right/left hip joint: denoted by Int.(+)/Ext.(-) Rot. (right/left).

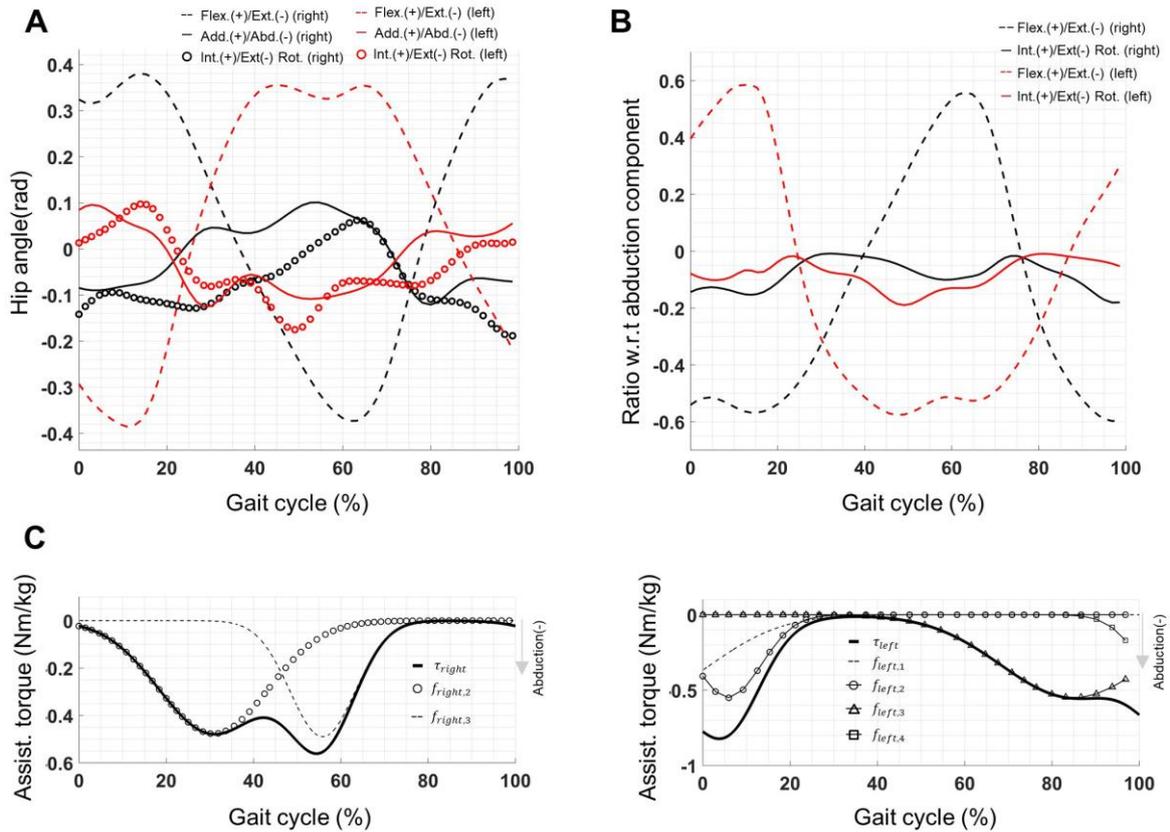

**Fig. S5. Sample gait motion and selected output moment profile by a modified R-4B-R mechanism from that synthesized for Case Study 1. (The modified geometric specification is given in Table S3B) (A)** Hip angles during the gait cycle of a sample gait motion, **(B)** the ratios of various moment components (e.g., the flexion/extension moment for the right hip joint, which is denoted by Flex.(+)/Ext.(-) (right)) with respect to the abduction moment components according to the sample gait motion, and **(C)** sample assistive torque profiles to be exerted on the actuator for the right hip mechanism (left) and for the left hip mechanism (right).

From the angle trajectory in Fig. S5A and the modified mechanism configuration in Table S3, we can obtain the output moment direction to be generated by the mechanism as plotted in Fig. S4B:

flexion/extension moment component with respect to the abduction component for the right/left hip joint: denoted by Flex.(+)/Ext.(-) (right/left), internal/external rotation moment component with respect to the abduction component for the right/left hip joint: denoted by Int.(+)/Ext.(-) Rot. (right/left).

In choosing the assistive torque profiles for the input actuator, we cite recent works (see, for instance, (46)), which select an input torque profile as the sum of the amplified normal distributions in the flexion/extension moment-assisting hip wearable robots. It is also well known that the distinctive double-peak pattern in the abduction hip moment profile is critical for stable gait (42). Based on these observations, we choose the input actuator torque profile as the superposition of two amplified normal distributions per one gait cycle. The explicit equations describing the input actuator torque profile are

$$\tau_{right} = \sum_{i=1}^{4} f_{right,i}, \tag{S31a}$$

$$\tau_{left} = \sum_{i=1}^{4} f_{left,i}, \tag{S31b}$$

where

$$f_{right,i}(x_{gait}) = \frac{\alpha_{right,i}}{\sigma_{right,i}\sqrt{2\pi}} \exp\left(-\frac{1}{2}\left(\frac{x_{gait} - \mu_{right,i}}{\sigma_{right,i}}\right)^2\right) \text{ (for } i = 1,2,3,4\text{)}, \tag{S32a}$$

$$f_{left,i}(x_{gait}) = \frac{\alpha_{left,i}}{\sigma_{left,i}\sqrt{2\pi}} \exp\left(-\frac{1}{2}\left(\frac{x_{gait} - \mu_{left,i}}{\sigma_{left,i}}\right)^2\right) \text{ (for } i = 1,2,3,4\text{)}. \tag{S32b}$$

In Eqs. S31a and S31b, $\tau_{right}$ and $\tau_{left}$ denote the torque profiles exerted on the actuators for right and left mechanisms, respectively. In Eqs. S32a and 32b, the symbol $x_{gait}$ represents a gait cycle in the units of percentage (or time). The values of the parameters appearing in equations 32a and 32b (such as $\sigma_{right,1}$, $\mu_{right,1}$, $\alpha_{right,1}$) are given in Table S4.

Table S4. Parameters used to define the input actuator profile shown in Fig. S5C

|  | Values in time domain (s) | Values in gait cycle domain (%) |
|---|---|---|
| $(\sigma_{right,1}, \mu_{right,1}, \alpha_{right,1})$ | (0.15, 0.9, -13) | (12.5, 30.83, -0.108) |
| $(\sigma_{right,2}, \mu_{right,2}, \alpha_{right,2})$ | (0.09, 1.2, -8) | (7.5, 55.83, -0.067) |
| $(\sigma_{right,3}, \mu_{right,3}, \alpha_{right,3})$ | (0.15, 2.1, -13) | (12.5, 130.83, -0.108) |
| $(\sigma_{right,4}, \mu_{right,4}, \alpha_{right,4})$ | (0.09, 2.4, -8) | (7.5, 155.83, -0.067) |
| $(\sigma_{left,1}, \mu_{left,1}, \alpha_{left,1})$ | (0.2, 1.55, -20) | (16.67, 15, -0.167) |
| $(\sigma_{left,2}, \mu_{left,2}, \alpha_{left,2})$ | (0.09, 1.83, -9) | (7.5, 5.83, -0.075) |
| $(\sigma_{left,3}, \mu_{left,3}, \alpha_{left,3})$ | (0.2, 0.35, -20) | (16.67, 85, -0.1667) |
| $(\sigma_{left,4}, \mu_{left,4}, \alpha_{left,4})$ | (0.09, 0.6, -9) | (7.5, 108.33, -0.075) |

The parameters in Table S4 were initially set in the time domain (s) and then transformed into the form suitable for the gait cycle (%). Note that 0.53s and 1.73s in the raw data in the time domain transform into 0% and 100% in the gait cycle domain, respectively. In Eq. S32a, $f_{right,2}$ and $f_{right,3}$ are two amplified normal distributions for the current gait cycle, while $f_{right,1}$ ($f_{right,4}$) is one of two distributions for the previous (next) cycle, whose peak point is closer to the current cycle. The other distribution components whose peak values are far from the current gait cycle are excluded from the actuator profile since their contributions to the current cycle are negligible. Because the percentage of gait cycle completion is based on the right hip joint, where the initial contact phase of the right leg is set to be 0%, the profile for the left hip joint should be appropriately adjusted. Accordingly, the torque profile for the left joint in equation S32b is constructed as the superposition of four normal distributions: two distributions $f_{left,1}$ and $f_{left,2}$ for a gait cycle and the other two ones $f_{left,3}$ and $f_{left,4}$ for the following gait cycle.

Since the primary objective of this study is to develop a computational synthesis method, the input torque profiles are not further tuned, which may be needed for actual robot control.

## S6. A 3D-printed prototype of the synthesized R-4B-R mechanism

Figure S6 shows an R-4B-R mechanism prototype. It was produced using the FDM (Fused Deposition Modeling) 3D printing. The connectivity of the mechanism is given in Figure 6, and the geometric details are provided in Table S3B. All mechanical components of the mechanism were meticulously fabricated because a spherical mechanism [2] is an overconstrained mechanism and can become a DOF-deficient configuration if not all axes of the revolute joints of the mechanism pass the common point precisely (the origin of the corresponding spherical coordinate system). Actual tests confirmed that the fabricated R-4B-R mechanism operates smoothly during the wearer's gait motion. The zoomed-in views in Fig. S6 depict the specific configurations of the mechanism when the right leg is in the flexed, neutral, and extended poses, in that order. The symbols R$i$ ($i = 1, \cdots$) and VR in the figure are the same symbols introduced in Fig. 6. As explained in the main text, the virtual revolute joint (VR) created by the spherical four-bar mechanism consisting of four revolute joints R2, R3, R4, and R5 and the revolute joint R6 moves back and forth to assist the moment in the programmed direction.

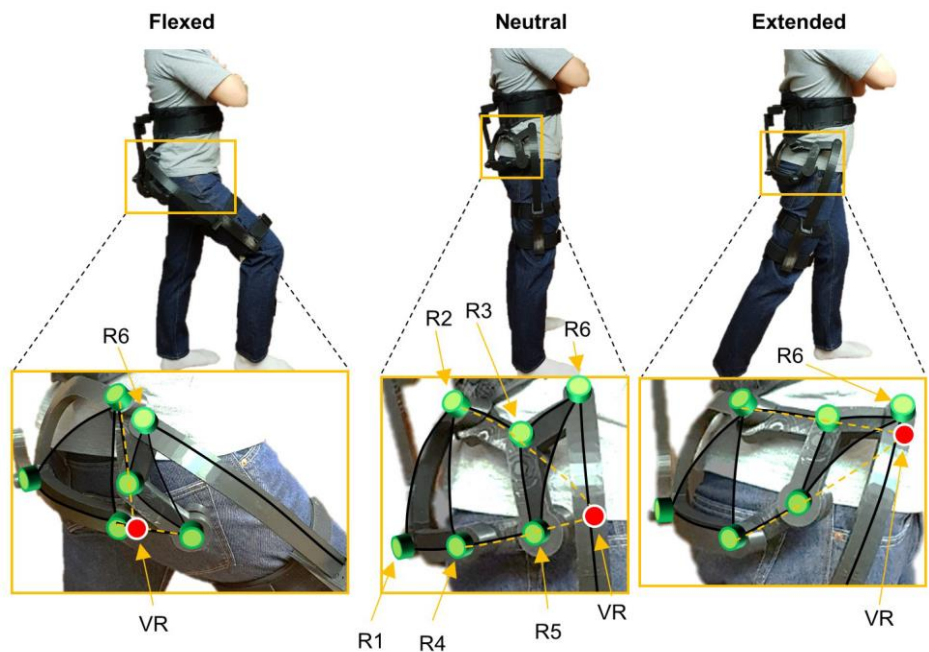

**Fig. S6. Snapshots of different poses of a human wearing the prototype of the synthesized R-4B-R mechanism (R1~R6: actual revolute joint, VR: virtual revolute joint; see also Fig. 6).**

---

[2] Recall that the synthesized R-4B-R mechanism is a spherical mechanism.